\documentclass[10pt,journal,compsoc]{IEEEtran}
\usepackage[
            pdfstartview=FitH,
            CJKbookmarks=true,
            bookmarksnumbered=true,
            bookmarksopen=true,
            colorlinks, 
            pdfborder=001,   
            linkcolor=green,
            anchorcolor=green,
            citecolor=green
            ]{hyperref}
  \usepackage[nocompress]{cite}

\usepackage{graphicx}
\usepackage{amsmath,amssymb} 
\usepackage{calligra}
\usepackage{frcursive}
\usepackage[T1]{fontenc}
\usepackage{caption}
\usepackage{color}
\usepackage{cite}
\usepackage{bm}
\usepackage{microtype}
\usepackage[tight,footnotesize]{subfigure}
\usepackage{multirow}
\usepackage{floatrow}
\usepackage{colortbl}
\usepackage{array}
\usepackage[figuresright]{rotating}
\definecolor{mygray}{gray}{.9}

\ifCLASSINFOpdf
\usepackage{bm}
\else
\fi
\newcommand{\tabincell}[2]{\begin{tabular}{@{}#1@{}}#2\end{tabular}}

\begin{document}

\title{SIFT Meets CNN: \\A Decade Survey of Instance Retrieval}

\author{Liang Zheng,
        Yi Yang,~
        and Qi Tian,~\IEEEmembership{Fellow, IEEE}

\IEEEcompsocitemizethanks{
\IEEEcompsocthanksitem L. Zheng and Y. Yang (corresponding author) are with the
Centre for AI, University of Technology at Sydney, NSW, Australia. \protect\\
E-mail: liang.zheng@uts.edu.au, yi.yang@uts.edu.au

\IEEEcompsocthanksitem  Q. Tian is with the Department of Computer Science, University of Texas at San Antonio, TX, 78256 USA. E-mail: qitian@cs.utsa.edu
}
}

\markboth{Journal of \LaTeX\ Class Files,~Vol.~14, No.~8, August~2015}%
{SYan2016hell \MakeLowercase{\textit{et al.}}: Bare Advanced Demo of IEEEtran.cls for IEEE Computer Society Journals}

\IEEEtitleabstractindextext{%
\begin{abstract}
In the early days, content-based image retrieval (CBIR) was studied with global features. Since 2003, image retrieval based on local descriptors (\emph{de facto} SIFT) has been extensively studied for over a decade due to the advantage of SIFT in dealing with image transformations. Recently, image representations based on the convolutional neural network (CNN) have attracted increasing interest in the community and demonstrated impressive performance. Given this time of rapid evolution, this article provides a comprehensive survey of instance retrieval over the last decade. Two broad categories, SIFT-based and CNN-based methods, are presented. For the former, according to the codebook size, we organize the literature into using large/medium-sized/small codebooks. For the latter, we discuss three lines of methods, \emph{i.e.,} using pre-trained or fine-tuned CNN models, and hybrid methods. The first two perform a single-pass of an image to the network, while the last category employs a patch-based feature extraction scheme. This survey presents milestones in modern instance retrieval, reviews a broad selection of previous works in different categories, and provides insights on the connection between SIFT and CNN-based methods. After analyzing and comparing retrieval performance of different categories on several datasets, we discuss promising directions towards generic and specialized instance retrieval.

\end{abstract}

\begin{IEEEkeywords}
Instance retrieval, SIFT, convolutional neural network, literature survey.
\end{IEEEkeywords}}

\maketitle

\IEEEdisplaynontitleabstractindextext

\IEEEpeerreviewmaketitle

\ifCLASSOPTIONcompsoc
\IEEEraisesectionheading{\section{Introduction}\label{sec:introduction}}
\else
\section{Introduction}
\label{sec:introduction}
\fi

\IEEEPARstart{C}{ontent}-based image retrieval (CBIR) has been a long-standing research topic in the computer vision society.
In the early 1990s, the study of CBIR truly started. Images were indexed by the visual cues, such as texture and color, and a myriad of algorithms and image retrieval systems have been proposed. A straightforward strategy is to extract global descriptors. This idea dominated the image retrieval community in the 1990s and early 2000s. Yet, a well-known problem is that global signatures may fail the invariance expectation to image changes such as illumination, translation, occlusion and truncation. These variances compromise the retrieval accuracy and limit the application scope of global descriptors. This problem has given rise to local feature based image retrieval.

The focus of this survey is instance-level image retrieval. In this task, given a query image depicting a particular object/scene/architecture, the aim is to retrieve images containing the same object/scene/architecture that may be captured under different views, illumination, or with occlusions. Instance retrieval departs from class retrieval \cite{torresani2010efficient} in that the latter aims at retrieving images of the same class with the query. In the following, if not specified, we use ``image retrieval'' and ``instance retrieval'' interchangeably.

\begin{figure*}[t]
  \centering
  \includegraphics[width=7.0in]{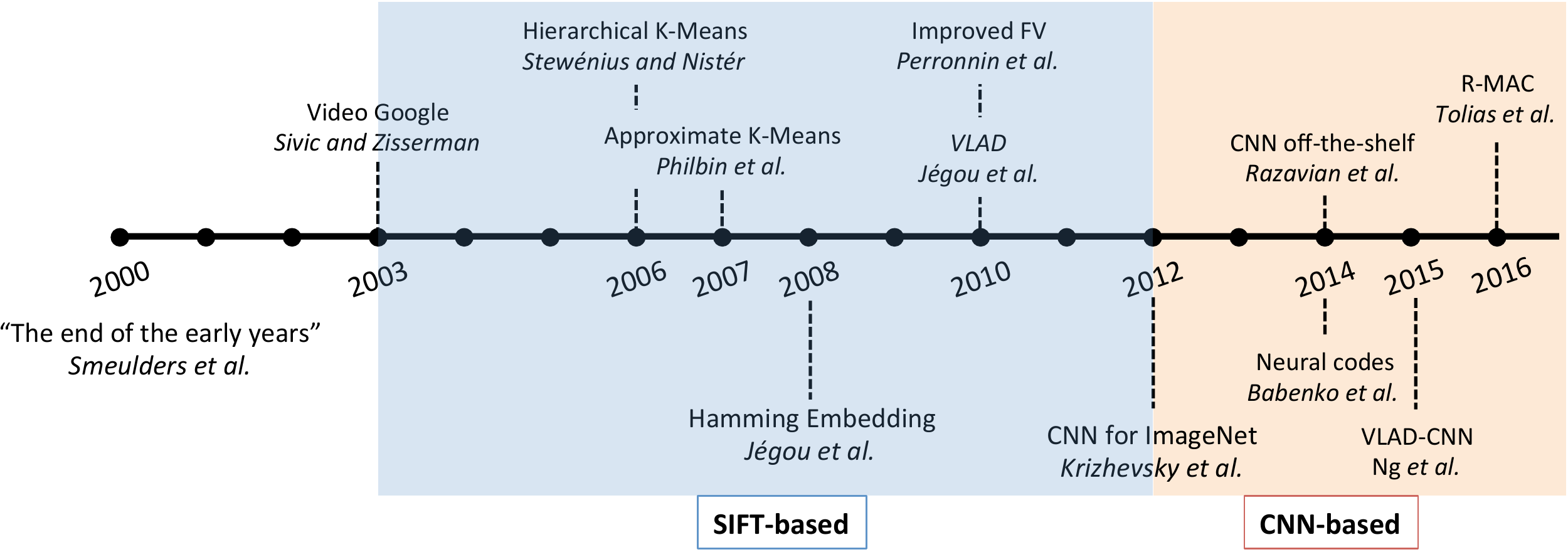}\\
  \caption{Milestones of instance retrieval. After a survey of methods before the year 2000 by Smeulders \emph{et al} \cite{smeulders2000content}, Sivic and Zisserman \cite{video_google} proposed Video Google in 2003, marking the beginning of the BoW model. Then, the hierarchical k-means and approximate k-means were proposed by Stew\'{e}nius and Nist\'{e}r \cite{HKM} and Philbin \emph{et al.} \cite{AKM}, respectively, marking the use of large codebooks in retrieval. In 2008, J\'{e}gou \emph{et al.} \cite{Hamming} proposed Hamming Embedding, a milestone in using medium-sized codebooks. Then, compact visual representations for retrieval were proposed by Perronnin \emph{et al.} \cite{perronnin2010improving} and J\'{e}gou \emph{et al.} \cite{jegou2010aggregating} in 2010. Although SIFT-based methods were still moving forward, CNN-based methods began to gradually take over, following the pioneering work of Krizhevsky \emph{et al.} \cite{krizhevsky2012imagenet}. In 2014, Razavian \emph{et al.} \cite{razavian2014cnn} proposed a hybrid method extracting multiple CNN features from an image. Babenko \emph{et al.} \cite{babenko2014neural} were the first to fine-tune a CNN model for generic instance retrieval. Both \cite{ng2015exploiting,tolias2016particular} employ the column features from pre-trained CNN models, and \cite{tolias2016particular} inspires later state-of-the-art methods. These milestones are the representative works of the categorization scheme in this survey. }\label{fig:timeline}
\end{figure*}

The milestones of instance retrieval in the past years are presented in Fig. \ref{fig:timeline}, in which the times of the SIFT-based and CNN-based methods are highlighted. The majority of traditional methods can be considered to end in 2000 when Smeulders \emph{et al.} \cite{smeulders2000content} presented a comprehensive survey of CBIR ``at the end of the early years''. Three years later (2003) the Bag-of-Words (BoW) model was introduced to the image retrieval community \cite{video_google}, and in 2004 was applied to image classification \cite{csurka2004visual}, both relying on the SIFT descriptor \cite{SIFT2}. The retrieval community has since witnessed the prominence of the BoW model for over a decade during which many improvements were proposed. In 2012, Krizhevsky \emph{et al.} \cite{krizhevsky2012imagenet} with the AlexNet achieved the state-of-the-art recognition accuracy in ILSRVC 2012, exceeding previous best results by a large margin. Since then, research focus has begun to transfer to deep learning based methods \cite{razavian2014cnn,babenko2014neural,ng2015exploiting,tolias2016particular}, especially the convolutional neural network (CNN).

The SIFT-based methods mostly rely on the BoW model. BoW was originally proposed for modeling documents because the text is naturally parsed into words. It builds a word histogram for a document by accumulating word responses into a global vector. In the image domain, the introduction of the scale-invariant feature transform (SIFT) \cite{SIFT2} makes the BoW model feasible \cite{video_google}. Originally, SIFT is comprised of a detector and descriptor, but which are used in isolation now;
in this survey, if not specified, SIFT usually refers to the 128-dim descriptor, a common practice in the community.
With a pre-trained codebook (vocabulary), local features are quantized to visual words. An image can thus be represented in a similar form to a document, and classic weighting and indexing schemes can be leveraged.


In recent years, the popularity of SIFT-based models seems to be overtaken by the convolutional neural network (CNN), a hierarchical structure that has been shown to outperform hand-crafted features in many vision tasks.
In retrieval, competitive performance compared to the BoW models has been reported, even with short CNN vectors \cite{tolias2016particular,kalantidis2016cross,gordo2016deep}. The CNN-based retrieval models usually compute compact representations and employ the Euclidean distance or some approximate nearest neighbor (ANN) search methods for retrieval. Current literature may directly employ the pre-trained CNN models or perform fine-tuning for specific retrieval tasks. A majority of these methods feed the image into the network only once to obtain the descriptor. Some are based on patches which are passed to the network multiple times, a similar manner to SIFT; we classify them into hybrid methods in this survey.

\subsection{Organization of This Paper}
Upon the time of change, this paper provides a comprehensive literature survey of both the SIFT-based and CNN-based instance retrieval methods. We first present the categorization methodology in Section \ref{sec:retrieval_type}. We then describe the two major method types in Section \ref{sec:sift_model} and Section \ref{sec:cnn_model}, respectively.
On several benchmark datasets, Section \ref{sec:experiment} summarizes the comparisons between SIFT- and CNN-based methods. In Section \ref{sec:directions}, we point out two possible future directions. This survey will be concluded in Section \ref{sec:conclusion}.

\begin{figure*}[t]
  \centering
  \includegraphics[width=7.0in]{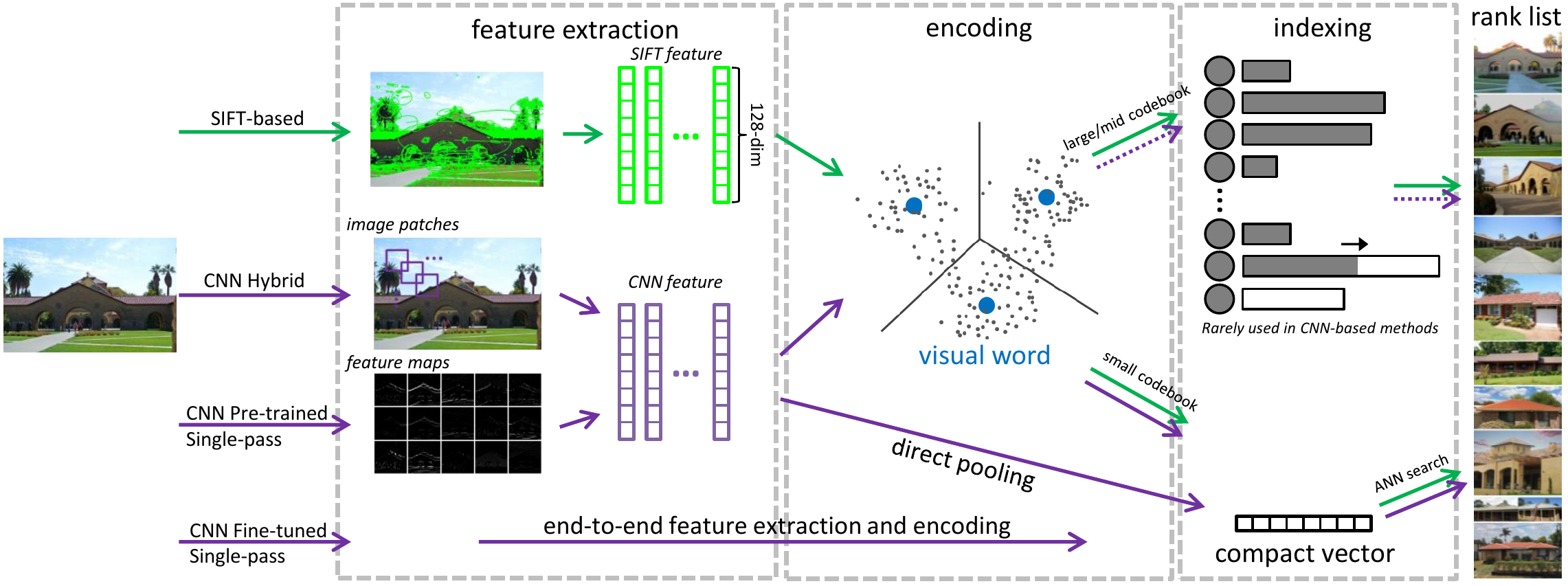}\\
  \caption{A general pipeline of SIFT- and CNN-based retrieval models. Features are computed from hand-crafted detectors for SIFT, and densely applied filters or image patches for CNN. In both methods, under small codebooks, encoding/pooling is employed to produce compact vectors. In SIFT-based methods, the inverted index is necessary under large/medium-sized codebooks. The CNN features can also be computed in an end-to-end way using fine-tuned CNN models. } \label{fig:pipeline}
\end{figure*}

\section{Categorization Methodology}\label{sec:retrieval_type}

\setlength{\tabcolsep}{4.4pt}
\begin{table*}[t]
\caption{Major differences between various types of instance retrieval models. For SIFT-based methods, hand-crafted local invariant features are extracted, and according to the codebook sizes, different encoding and indexing strategies are leveraged. For CNN-based methods, pre-trained, fine-tuned CNN models and hybrid methods are the primary types; fixed-length compact vectors are usually produced, combined with approximate nearest neighbor (ANN) methods.  }
\small
\footnotesize
\renewcommand{\arraystretch}{1.1}
\centering
\begin{tabular}{l |l |l| l |l | l|l  }
\hline
\multicolumn{2}{c|}{method type} & detector &  descriptor& encoding&dim. & indexing \\
\hline
\multirow{3}{*}{SIFT-based} &Large voc.& \multirow{3}{*}{\tabincell{l}{DoG, Hessian-Affine,\\ dense patches, \emph{etc}.}} &\multirow{3}{*}{\tabincell{l}{Local invariant descriptors\\ such as SIFT}}&Hard, soft&High& Inverted index  \\
 \cline{2-2}
\cline{5-7}
 &Mid voc.&  &&Hard, soft, HE&Medium& Inverted index   \\
 \cline{2-2}
  \cline{5-7}
 &Small voc.&  &&VLAD, FV&Low& ANN methods   \\
\hline
 \multirow{3}{*}{CNN-based}&Hybrid& Image patches& CNN features & VLAD, FV, pooling & Varies & ANN methods  \\
\cline{2-7}
&Pre-trained, single-pass& \multicolumn{2}{l|}{Column feat. or FC of pre-trained CNN models.}& VLAD, FV, pooling & Low  &ANN methods \\
\cline{2-7}
 &Fine-tuned, single-pass& \multicolumn{3}{l|}{A global feat. is end-to-end extracted from fine-tuned CNN models.} & Low & ANN methods  \\
\hline
\end{tabular}
\label{table:three_types_methods}
\end{table*}

According to the different visual representations, this survey categorizes the retrieval literature into two broad types: SIFT-based and CNN-based. The SIFT-based methods are further organized into three classes: using large, medium-sized or small codebooks. We note that the codebook size is closely related to the choice of encoding methods. The CNN-based methods are categorized into using pre-trained or fine-tuned CNN models, as well as hybrid methods. Their similarities and differences are summarized in Table \ref{table:three_types_methods}.

\textbf{The SIFT-based methods} had been predominantly studied before 2012 \cite{krizhevsky2012imagenet} (good works also appear in recent years \cite{jegou2014triangulation,shi2015early}). This line of methods usually use one type of detector, \emph{e.g.,} Hessian-Affine, and one type of descriptor, \emph{e.g.,} SIFT. Encoding maps a local feature into a vector. Based on the size of the codebook used during encoding, we classify SIFT-based methods into three categories as below.
\begin{itemize}[leftmargin=0pt]
  \item \textbf{Using small codebooks.} The visual words are fewer than several thousand. Compact vectors are generated \cite{jegou2010aggregating,perronnin2010improving} before dimension reduction and coding.
  \item \textbf{Using medium-sized codebooks.} Given the sparsity of BoW and the low discriminative ability of visual words, the inverted index and binary signatures are used \cite{Hamming}. The trade-off between accuracy and efficiency  is a major influencing factor \cite{selective_match}.
  \item \textbf{Using large codebooks.} Given the sparse BoW histograms and the high discriminative ability of visual words, the inverted index and memory-friendly signatures are used \cite{GVP}. Approximate methods are used in codebook generation and encoding \cite{HKM,AKM}.
\end{itemize}

\textbf{The CNN-based methods} extract features using CNN models. Compact (fixed-length) representations are usually built. There are three classes:
\begin{itemize}
  \item \textbf{Hybrid methods.} Image patches are fed into CNN multiple times for feature extraction \cite{razavian2014cnn}. Encoding and indexing are similar to SIFT-based methods \cite{gong2014multi}.
  \item \textbf{Using pre-trained CNN models.} Features are extracted in a single pass using CNN pre-trained on some large-scale datasets like ImageNet \cite{deng2009imagenet}. Compact Encoding/pooling techniques are used \cite{ng2015exploiting,tolias2016particular}.
  \item \textbf{Using fine-tuned CNN models.} The CNN model (\emph{e.g.,} pre-trained on ImageNet) is fine-tuned on a training set in which the images share similar distributions with the target database \cite{babenko2014neural}. CNN features can be extracted in an end-to-end manner through a single pass to the CNN model. The visual representations exhibit improved discriminative ability \cite{radenovic2016cnn,gordo2016deep}.
\end{itemize}



\section{SIFT-based Image Retrieval}\label{sec:sift_model}
\subsection{Pipeline} \label{sec:sift_pipeline}
The pipeline of SIFT-based retrieval is introduced in Fig. \ref{fig:pipeline}.

\textbf{Local feature extraction.} Suppose we have a gallery $\mathcal{G}$ consisting of $N$ images. Given a feature detector, we extract local descriptors from the regions around the sparse interest points or dense patches. We denote the local descriptors of $D$ detected regions in an image as $\{f_i\}_{i = i}^D, f_i\in \mathbb{R}^p$.


\textbf{Codebook training.} SIFT-based methods train a codebook offline. 
Each visual word in the codebook lies in the center of a subspace, called the ``Voronoi cell''. A larger codebook corresponds to a finer partitioning, resulting in more discriminative visual words and vice versa.
Suppose that a pool of local descriptors $\mathcal{F} \equiv \{f_i\}_{i=1}^M$ are computed from an unlabeled training set. The baseline approach, \emph{i.e.,} k-means, partitions the $M$ points into $K$ clusters; the $K$ visual words thus constitute a codebook of size $K$.


\textbf{Feature encoding.}  A local descriptor $f_i \in \mathbb{R}^p$ is mapped into a feature embedding $g_i  \in \mathbb{R}^l$ through the feature encoding process, $f_i \rightarrow g_i$.
%
When k-means clustering is used, $f_i$ can be encoded according to its distances to the visual words. 
For large codebooks, hard \cite{AKM,HKM} and soft quantization \cite{soft} are good choices. In the former, the resulting embedding $g_i$ has only one non-zero entry; in the latter, $f_i$ can be quantized to a small number of visual words. A global signature is produced after a sum-pooling of all the embeddings of local features.
For medium-sized codebooks, additional binary signatures can be generated to preserve the original information. When using small codebooks, popular encoding schemes include vector of locally aggregated descriptors (VLAD)  \cite{jegou2010aggregating}, Fisher vector (FV) \cite{perronnin2010improving}, \emph{etc}.



\subsection{Local Feature Extraction}\label{sec:local_feature_extraction}
Local invariant features aim at accurate matching of local structures between images \cite{tuytelaars2008local}.
SIFT-based  methods usually share a similar feature extraction step composed of a feature detector and a descriptor.

\textbf{Local detector.} The \textbf{interest point detectors} aim to reliably localize a set of stable local regions under various imaging conditions. 
In the retrieval community, finding affine-covariant regions has been preferred. It is called ``covariant'' because the shapes of the detected regions change with the affine transformations, so that the region content (descriptors) can be invariant. This kind of detectors are different from keypoint-centric detectors such as the Hessian detector \cite{beaudet1978rotationally}, and from those focusing on scale-invariant regions such as the difference of Gaussians (DoG) \cite{SIFT2} detector. Elliptical regions which are adapted to the local intensity patterns are produced by affine detectors. This ensures that the same local structure is covered under deformations caused by viewpoint variances, a problem often encountered in instance retrieval. 
In the  milestone work \cite{video_google}, the Maximally Stable Extremal Region (MSER) detector \cite{matas2004robust} and the affine extended Harris-Laplace detector are employed, both of which are affine-invariant region detectors. MSER is  used in several later works \cite{wu2009bundling,HKM}. 
Starting from \cite{AKM}, the Hessian-affine detector \cite{mikolajczyk2005comparison} has been widely adopted in retrieval. It has been shown to be superior to the difference of Gaussians (DoG) detector \cite{Hamming, zheng2014packing}, due to its advantage in reliably detecting local structures under large viewpoint changes. To fix the orientation ambiguity of these affine-covariant regions, the gravity assumption is made \cite{perd2009efficient}. The practice which dismisses the orientation estimation is employed by later works \cite{root_sift,simonyan2014learning} and demonstrates consistent improvement on architecture datasets where the objects are usually upright. Other non-affine detectors have also been tested in retrieval, such as the Laplacian of Gaussian (LOG) and Harris detectors used in  \cite{ge2013sparse}. For objects with smooth surfaces \cite{arandjelovic2011smooth}, few interest points can be detected, so the object boundaries are good candidates for local description.

On the other hand, some employ the \textbf{dense region detectors}. In the comparison between densely sampled image patches and the detected patches, Sicre \emph{et al.} \cite{sicre2014dense} report the superiority of the former. To recover the rotation invariance of dense sampling, the dominant angle of  patches is estimated in \cite{zhao2013oriented}. A comprehensive comparison of various dense sampling strategies, the interest point detectors, and those in between can be accessed in  \cite{iscen2015comparison}.

\textbf{Local Descriptor.} With a set of detected regions, descriptors encode the local content. SIFT \cite{SIFT2} has been used as the default descriptor. The 128-dim vector has been shown to outperform competing descriptors in matching accuracy \cite{mikolajczyk2005performance}. In an extension, PCA-SIFT \cite{ke2004pca} reduces the dimension from 128 to 36 to speed up the matching process at the cost of more time in feature computation and loss of distinctiveness. Another improvement is RootSIFT \cite{root_sift}, calculated by two steps: 1) $\ell_1$ normalize the SIFT descriptor, 2) square root each element. 
RootSIFT is now used as a routine in SIFT-based retrieval. Apart from SIFT, SURF \cite{bay2006surf} is also widely used. It combines the Hessian-Laplace detector and a local descriptor of the local gradient histograms. The integral image is used for acceleration. SURF has a comparable matching accuracy with SIFT and is faster to compute. See \cite{juan2009comparison} for comparisons between SIFT, PCA-SIFT, and SURF. To further accelerate the matching speed, binary descriptors  \cite{rublee2011orb} replace  Euclidean distance with Hamming distance during matching. 

Apart from hand-crafted descriptors, some also propose learning schemes to improve the discriminative ability of local descriptors. For example, Philbin \emph{et al.} \cite{philbin2010descriptor} proposes a non-linear transformation so that the projected SIFT descriptor yields smaller distances for true matches. 
Simoyan \emph{et al.} \cite{simonyan2014learning} improve this process by learning both the pooling region and a linear descriptor projection.




\subsection{Retrieval Using Small Codebooks}\label{sec:small_codebook}
A small codebook has several thousand, several hundred or fewer visual words, so the computational complexity of codebook generation and encoding is moderate. Representative works  include BoW \cite{video_google}, VLAD \cite{jegou2010aggregating} and FV \cite{perronnin2010improving}. We mainly discuss VLAD and FV and refer readers to \cite{radenovic2015multiple} for a comprehensive evaluation of the BoW compact vectors.

\subsubsection{Codebook Generation}



Clustering complexity depends heavily on the codebook size. In works based on VLAD \cite{jegou2010aggregating} or FV \cite{perronnin2010improving}, the codebook sizes are typically small, \emph{e.g.,} 64, 128, 256. For VLAD, flat k-means is employed for codebook generation. For FV, the Gaussian mixture model (GMM), \emph{i.e.,} $u_{\lambda}(x) = \sum_{i=1}^K w_i u_i(x)$, where $K$ is the number of Gaussian mixtures, is trained using the maximum likelihood estimation. GMM describes the feature space with a mixture of $K$ Gaussian distributions, and can be denoted as $\lambda = \{w_i, \mu_i, \sum_i, i = 1,...,K\}$, where $w_i$, $\mu_i$ and $\sum_i$ represent the mixture weight, the mean vector and the covariance matrix of Gaussian $u_i$, respectively.

\subsubsection{Encoding}\label{sec:encoding_small}
Due to the small codebook size, relative complex and information-preserving encoding techniques can be applied. We mainly describe FV, VLAD and their improvements in this section. 
With a pre-trained GMM model, FV describes the averaged first and second order difference between local features and the GMM centers. Its dimension is $2pK$, where $p$ is the dimension of the local descriptors and $K$ is the codebook size of GMM. FV usually undergoes power normalization \cite{perronnin2010large}, \cite{perronnin2010improving} to suppress the burstiness problem (to be described in Section \ref{sec:tf-idf}). In this step, each component of FV undergoes non-linear transformation featured by parameter $\alpha$, $x_i := \mbox{sign}(x_i)\|x_i\|^\alpha$. Then $\ell_2$ normalization is employed.
Later, FV is improved from different aspects. For example, Koniusz \emph{et al.} \cite{koniusz2013comparison} augment each descriptor with its spatial coordinates and associated tunable weights. In \cite{gosselin2014revisiting}, larger codebooks (up to 4,096) are generated and demonstrate superior classification accuracy to smaller codebooks, at the cost of computational efficiency. To correct the assumption that local regions are identically and independently distributed (iid), Cinbis \emph{et al.} \cite{cinbis2015approximate} propose non-iid models that discount the burstiness effect and yield improvement over the power normalization.

The VLAD encoding scheme proposed by J\'{e}gou \emph{et al.} \cite{jegou2010aggregating} can be thought of as a simplified version of FV. It quantizes a local feature to its nearest visual word in the codebook and records the difference between them.  Nearest neighbor search is performed because of the small codebook size. The residual vectors are then aggregated by sum pooling followed by normalizations. The dimension of VLAD is $pK$. Comparisons of some important encoding techniques are presented in \cite{chatfield2011devil,jegou2012aggregating}.  
Again, the improvement of VLAD comes from multiple aspects. In \cite{jegou2012negative}, J\'{e}gou and Chum suggest the usage of PCA and whitening (denoted as PCA$_w$ in Table \ref{table:compare_methods_datasets}) to de-correlate visual word co-occurrences, and the training of multiple codebooks to reduce quantization loss. In \cite{arandjelovic2013all}, Arandjelovi{\'c} \emph{et al.} extend  VLAD in three aspects: 1) normalize the residual sum within each coarse cluster, called intra-normalization, 2) vocabulary adaptation to address the dataset transfer problem and 3) multi-VLAD for small object discovery. Concurrent to \cite{arandjelovic2013all}, Delhumeau \emph{et al.} \cite{delhumeau2013revisiting} propose to normalize each residual vector instead of the residual sums; they also advocate for local PCA within each Voronoi cell which does not perform dimension reduction as \cite{jegou2012aggregating}. 
A recent work \cite{husain2016improving} employs soft assignment and empirically learns optimal weights for each rank to improve over the hard quantization. 

Note that some general techniques benefit various embedding methods, such as VLAD, FV, BoW, locality-constrained linear coding (LLC) \cite{wang2010locality} and monomial embeddings. To improve the discriminative ability of embeddings, Tolias \emph{et al.} \cite{tolias2014orientation} propose the orientation covariant embedding to encode the dominant orientation of the SIFT regions jointly with the SIFT descriptor. It achieves a similar covariance property to weak geometric consistency (WGC) \cite{Hamming} by using geometric cues within regions of interest so that matching points with similar dominant orientations are up-weighted and vice versa.  The triangulation embedding \cite{jegou2014triangulation} only considers the direction instead of the magnitude of the input vectors. 
J\'{e}gou \emph{et al.} \cite{jegou2014triangulation} also present a democratic aggregation that limits the interference between the mapped vectors. Baring a similar idea with democratic aggregation, Murray and Perronnin \cite{murray2014generalized} propose the generalized max pooling (GMP) optimized by equalizing the similarity between the pooled vector and each coding representation. 

The computational complexity of BoW, VLAD and FV is similar. We neglect the offline training and SIFT extraction steps. During visual word assignment, each feature should compute its distance (or soft assignment coefficient) with all the visual words (or Gaussians) for VLAD (or FV). So this step has a complexity of $\mathcal{O}(pK)$. In the other steps, complexity does not exceed $\mathcal{O}(pK)$. Considering the sum-pooling of the embeddings, the encoding process has an overall complexity of $\mathcal{O}(pKD)$, where $D$ is the number of features in an image. Triangulation embedding \cite{jegou2014triangulation}, a variant of VLAD, has a similar complexity. The complexity of multi-VLAD \cite{arandjelovic2013all} is $\mathcal{O}(pKD)$, too, but it has a more costly matching process. 
Hierarchical VLAD \cite{liu2016fine} has a complexity of $\mathcal{O}(pKK'D)$, where $K'$ is the size of the secondary codebook. In the aggregation stage, both GMP \cite{murray2014generalized}  and democratic aggregation \cite{jegou2014triangulation} have high complexity. The complexity of GMP  is $\mathcal{O}(\frac{P^2}{K})$, where $P$ is the dimension of the feature embedding, while the computational cost of democratic aggregation comes from the Sinkhorn algorithm.

\subsubsection{ANN Search}
Due to the high dimensionality of the VLAD/FV embeddings, efficient compression and ANN search methods have been employed \cite{jegou2011product,muja2014scalable}. For example, the principle component analysis (PCA) is usually adapted to for dimension reduction, and it is shown that retrieval accuracy even increases after PCA \cite{jegou2012negative}. For hashing-based ANN methods, Perronnin \emph{et al.} \cite{perronnin2010large} use standard binary encoding techniques such as locality sensitive hashing \cite{indyk1998approximate} and spectral hashing \cite{weiss2009spectral}. Nevertheless, when being tested on the SIFT and GIST feature datasets, spectral hashing is shown to be outperformed by Product Quantization (PQ) \cite{jegou2011product}. In these quantization-based ANN methods, PQ is demonstrated to be better than other popular ANN methods such as FLANN \cite{muja2014scalable} as well. A detailed discussion of VLAD and PQ can be viewed in \cite{spyromitros2014comprehensive}. PQ has since then been improved in a number of works. In \cite{douze2016polysemous}, Douze \emph{et al.} propose to re-order the cluster centroids so that adjacent centroids have small Hamming distances. This method is compatible with Hamming distance based ANN search, which offers significant speedup for PQ. We refer readers to \cite{wang2016survey} for a survey of ANN approaches.

We also mention an emerging ANN technique, \emph{i.e.,} group testing \cite{shi2014group}, \cite{iscen2016efficient}, \cite{iscen2017memory}. In a nutshell, the database is decomposed into groups, each represented by a group vector. Comparisons between the query and group vectors reveal how likely a group contains a true match. Since group vectors are much fewer than the database vectors, search time is reduced. Iscen \emph{et al.} \cite{iscen2016efficient} propose to directly find the best group vectors summarizing the database without explicitly forming the groups, which reduces the memory consumption.

\makeatother
\begin{figure} [t]
\centering
\subfigure[HKM]{\label{fig:nCam}%
\includegraphics[width=1.7in]{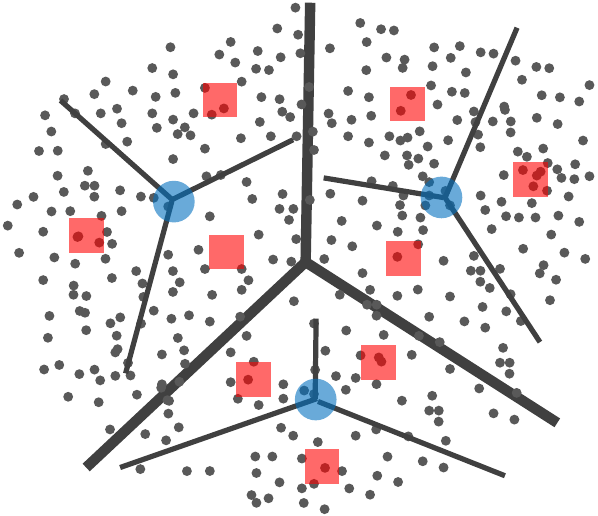}}
\hspace{0.0in}
 \subfigure[AKM]{\label{fig:nFrame}%
\includegraphics[width=1.71in]{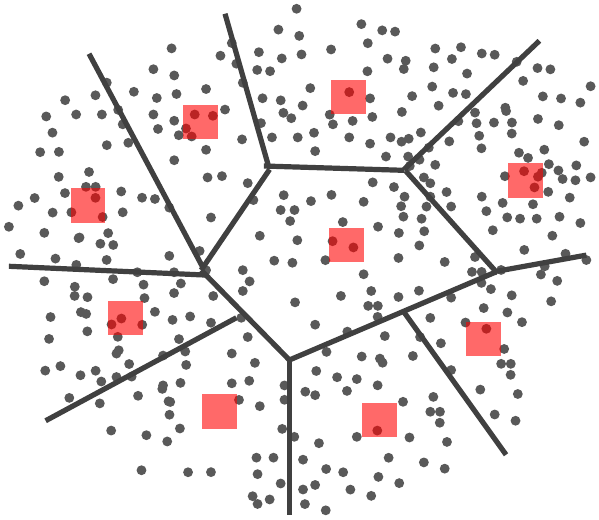}}
\caption{Two milestone clustering methods (a) hierarchical k-means (HKM) \cite{HKM} and (b) approximate k-means (AKM) \cite{AKM} for large codebook generation. Bold borders and blue discs are the clustering boundaries and centers of the first layer of HKM. Slim borders and red squares are the final clustering results in both methods.}
\label{fig:clustering}
\end {figure}
\subsection{Retrieval Using Large Codebooks}\label{sec:large_codebook}
A large codebook may contain 1 million \cite{AKM,HKM} visual words or more \cite{mikulik2010learning, co_indexing}. Some major steps undergo important changes compared with using small codebooks.
\subsubsection{Codebook Generation}\label{sec:codebook_large}
Approximate methods are critical in assigning data into a large number of clusters. In the retrieval community, two representative works are hierarchical k-means (HKM) \cite{HKM} and approximate k-means (AKM) \cite{AKM}, as illustrated in Fig. \ref{fig:timeline} and Fig. \ref{fig:clustering}. Proposed in 2006, HKM applies standard k-means on the training features hierarchically. It first partitions the points into a few clusters (\emph{e.g.,} $\bar{k} \ll K$) and then recursively partitions each cluster into further clusters. In every recursion, each point should be assigned to one of the $\bar{k}$ clusters, with the depth of the cluster tree being $\mathcal{O}(\log K)$, where $K$ is the target cluster number. The computational cost of HKM is therefore $\mathcal{O}(\bar{k}M\log K)$, where $M$ is the number of training samples. It is much smaller than the complexity of flat k-means $\mathcal{O}(MK)$ when $K$ is large (a large codebook).

The other milestone in large codebook generation is AKM \cite{AKM}. This method indexes the $K$ cluster centers using a forest of random $k$-d trees so that the assignment step can be performed efficiently with ANN search. In AKM, the cost of assignment can be written as $\mathcal{O}(K\log K + vM\log K) = \mathcal{O}(vM\log K)$, where $v$ is the number of nearest cluster candidates to be accessed in the $k$-d trees. So the computational complexity of AKM is on par with HKM and is significantly smaller than flat k-means when $K$ is large. Experiments show that AKM is superior to HKM \cite{AKM} due to its lower quantization error (see Section \ref{sec:quantization_large}). In most AKM-based methods, the default choice for ANN search is FLANN \cite{muja2014scalable}.




\subsubsection{Feature Encoding (Quantization)}\label{sec:quantization_large}
Feature encoding is interleaved with codebook clustering, because ANN search is critical in both components. The ANN techniques implied in some classic methods like AKM and HKM can be used in both clustering and encoding steps. Under a large codebook, the key trade-off is between quantization error and computational complexity.
In the encoding step, information-preserving encoding methods such as FV \cite{perronnin2010improving}, sparse coding \cite{yang2009linear} are mostly infeasible due to their computational complexity. It therefore remains a challenging problem how to reduce the quantization error while keeping the quantization process efficient.

Fro the ANN methods, 
the earliest solution is to quantize a local feature along the hierarchical tree structure \cite{HKM}. Quantized tree nodes in different levels are assigned different weights. However, due to the highly imbalanced tree structure, this method is outperformed by $k$-d tree based quantization method \cite{AKM}: one visual word is assigned to each local feature, using a $k$-d tree built from the codebook for fast ANN search. In an improvement to this hard quantization scheme, Philbin \emph{et al.} \cite{soft} propose soft quantization by quantizing a feature into several nearest visual words. The weight of each assigned visual word relates negatively to its distance from the feature by $\exp (-\frac{d^2}{2\sigma^2})$, where $d$ is the distance between the descriptor and the cluster center.
While  soft quantization is based on the Euclidean distance, Mikulik \emph{et al.} \cite{mikulik2010learning} propose to find relevant visual words for each visual word through an unsupervised set of matching features. Built on a probabilistic model, these alternative words tend to contain descriptors of matching features.
To reduce the memory cost of soft quantization \cite{soft} and the number of query visual words, Cai \emph{et al.} \cite{cai2012constrained} suggest that when a local feature is far away from even the nearest visual word, this feature can be discarded without a performance drop. To further accelerate  quantization, scalar quantization \cite{zhou2012scalar} suggests that local features be quantized without an explicitly trained codebook. A floating-point vector is binarized, and the first dimensions of the resulting binary vector are directly converted to a decimal number as a visual word. In the case of large quantization error and low recall, scalar quantization uses bit-flop to generate hundreds of visual words for a local feature. 

\subsubsection{Feature Weighting}\label{sec:tf-idf}
\textbf{TF-IDF.}
The visual words in codebook $\mathcal{C}$ are typically assigned specific weights, called the term frequency and inverse document frequency (TF-IDF), which are integrated with the BoW encoding. TF is defined as:
\begin{equation}\label{eq:tf}
\mbox{TF}(c_i^j) = o_i^j,
\end{equation}
where $o_i^j$ is the number of occurrences of a visual word $c_i$ within an image $j$. TF is thus a local weight.
IDF, on the other hand, determines the contribution of a given visual word through global statistics. The classic IDF weight of visual word $c_i$ is calculated as:
\begin{equation}\label{eq:idf}
\mbox{IDF}(c_i) = \log{\frac{N}{n_i}}, \mbox{where~} n_i = \sum_{j\in \mathcal{G}} \textbf{1}(o_i^j > 0),
\end{equation}
where $N$ is the number of gallery images, and $n_i$ encodes the number of images in which word $c_i$ appears. The TF-IDF weight for visual word $c_i$ in image $j$ is:
\begin{equation}\label{eq:weight}
w(c_i^j) = \mbox{TF}(c_i^j)\mbox{IDF}(c_i).
\end{equation}

\textbf{Improvements.}
A major problem associated with visual word weighting is burstiness \cite{burstiness}. It refers to the phenomenon whereby repetitive structures appear in an image. This problem tends to dominate image similarity. J{\'e}gou \emph{et al.} \cite{burstiness} propose several TF variants to deal with burstiness. An effective strategy consists in exerting a square operation on TF. Instead of grouping features with the same word index, Revaud \emph{et al.} \cite{revaud2012correlation} propose detecting keypoint groups frequently happening in irrelevant images which are down-weighted in the scoring function. While the above two methods detect bursty groups after quantization, Shi \emph{et al} \cite{shi2015early} propose detecting them in the descriptor stage. The detected bursty descriptors undergo average pooling and are fed in the BoW architectures. From the aspect of IDF, Zheng \emph{et al.} \cite{zheng2013lp} propose the $\mathcal{L}_p$-norm IDF to tackle burstiness and Murata \emph{et al.} \cite{murata2014bm25} design the exponential IDF which is later incorporated into the BM25 formula. When most works try to suppress burstiness, Torii \emph{et al} \cite{torii2013visual} view it as a distinguishing feature for architectures and design new similarity measurement following burstiness detection.

Another feature weighting strategy is feature augmentation on the database side \cite{turcot2009better}, \cite{root_sift}. Both methods construct an image graph offline, with edges indicating whether two images share a same object. For \cite{turcot2009better}, only features that pass the geometric verification are preserved, which reduces the memory cost. Then, the feature of the base image is augmented with all the visual words of its connecting images. This method is improved in \cite{root_sift} by only adding those visual words which are estimated to be visible in the augmented image, so that noisy visual words can be excluded.

\subsubsection{The Inverted Index}
\begin{figure}[t]
  \centering
  \includegraphics[width=3.4in]{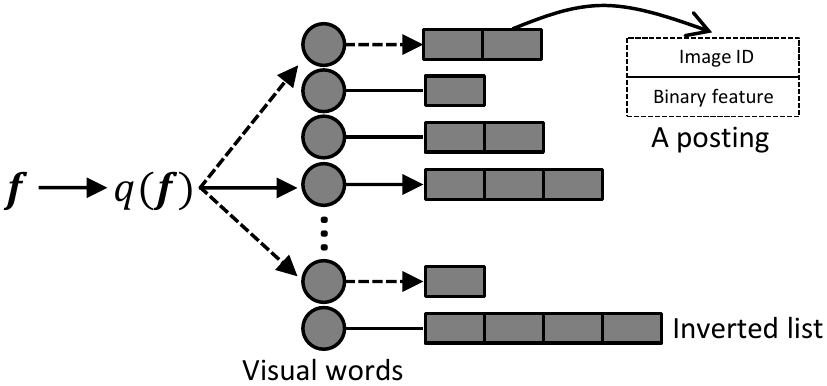}\\
  \caption{The data structure of the inverted index. It physically contains $K$ inverted lists, each consisting of some postings, which index the image ID and some binary signatures. During retrieval, a quantized feature will traverse the inverted list corresponding to its assigned visual word. Dashed line denotes soft quantization, in which multiple inverted lists are visited. }\label{fig:inv_index}
\end{figure}

The inverted index is designed to enable efficient storage and retrieval and is usually used under large/medium-sized codebooks. Its structure is illustrated in Fig. \ref{fig:inv_index}.
The inverted index is a one-dimensional structure where each entry corresponds to a visual word in the codebook. An inverted list is attached to each word entry, and those indexed in the each inverted list are called indexed features or postings. The inverted index takes advantages of the sparse nature of the visual word histogram under a large codebook. 

In literature, it is required that new retrieval methods be adjustable to the inverted index. In the baseline \cite{AKM, HKM}, the image ID and term frequency (TF) are stored in a posting. When other information is integrated, they should be small in size. For example, in \cite{contextual_weighting}, the metadata are quantized, such as descriptor contextual weight, descriptor density, mean relative log scale and the mean orientation difference in each posting. Similarly, quantized spatial information such as the orientation can also be stored \cite{sparse_coding}, \cite{GVP}. 
In co-indexing \cite{co_indexing}, when the inverted index is enlarged with globally consistent neighbors, semantically isolated images are deleted to reduce memory consumption. In \cite{babenko2012inverted}, the original one-dimensional inverted index is expanded to two-dimensional for ANN search, which learns a codebook for each SIFT sub-vector. Later, it is applied to instance retrieval by \cite{zheng2014packing} to fuse local color and SIFT descriptors. 

\subsection{Retrieval Using  Medium-sized Codebooks}\label{sec:medium_codebook}
Medium-sized codebooks refer to those having 10-200k visual words. 
The visual words  exhibit medium discriminative ability, and the inverted index is usually constructed.

\subsubsection{Codebook Generation and Quantization}
Considering the relatively small computational cost compared with large codebooks (Section \ref{sec:codebook_large}), flat k-means can be adopted for codebook generation \cite{jegou2010improving}, \cite{selective_match}. It is also shown in \cite{zheng2014packing,zheng2015query} that using AKM \cite{AKM} for clustering also yields very competitive retrieval accuracy.

For quantization, nearest neighbor search can be used to find the nearest visual words in the codebook. Practice may tell that using some strict ANN algorithms produces competitive retrieval results. So comparing with the extensive study on quantization under large codebooks (Section \ref{sec:quantization_large}) \cite{soft,mikulik2010learning,cai2012constrained}, relatively fewer works focus on the quantization problem under a medium-sized codebook.

\subsubsection{Hamming Embedding and its improvements}
The discriminative ability of visual words in medium-sized codebooks lies in between that of  small and large codebooks. So it is important to compensate the information loss during quantization. To this end, a milestone work, \emph{i.e.,} Hamming embedding (HE) has been dominantly employed.

Proposed by J{\'e}gou \emph{et al.} \cite{Hamming}, HE greatly improves the discriminative ability of visual words under medium-sized codebooks. HE first maps a SIFT descriptor $f\in \mathbb{R}^p$ from the $p$-dimensional  space to a $p_b$-dimensional space:
\begin{equation}
x = P\cdot f = (x_1,...x_{p_b}),
\end{equation}
where $P\in \mathbb{R}^p_b\times p$ is a projecting matrix, and $x$ is a low-dimensional vector. By creating a matrix of random Gaussian values and applying a QR factorization to it,  matrix $P$ is taken as the first $p_b$ rows of the resulting orthogonal matrix. To binarize $x$, Jegou \emph{et al.} propose to compute the median vector $\overline{x_i} = (\overline{x_{1,i}},...,\overline{x_{p_b, i}})$ of the low-dimensional vector using descriptors falling in each Voronoi cell $c_i$. Given descriptor $f$ and its projected vector $x$, HE computes its visual word $c_t$, and the HE binary vector is computed as:
\begin{equation}
b_j(x) =
\begin{cases}
1 \mbox{~~if~} x_j > \overline{x_{j, t}},\\
0 \mbox{~~otherwise}
\end{cases},
\end{equation}
where $b(x) = (b_1(x),...,b_{p_b}(x))$ is the resulting HE vector of dimension $p_b$. The binary feature $b(x)$ serves as a secondary check for feature matching. A pair of local features are a true match when two criteria are satisfied: 1) identical visual words and 2) small Hamming distance between their HE signatures. The extension of HE \cite{jegou2010improving} estimates the matching strength between feature $f_1$ and $f_2$ reversely to the Hamming distance by an exponential function:
\begin{equation}
w_{\mbox{HE}}(f_1, f_2) = \exp(-\frac{\mathcal{H}(b(x_1), b(x_2))}{2\gamma^2}),
\end{equation}
where $b(x_1)$ and $b(x_2)$ are the HE binary vector of $f_1$ and $f_2$, respectively, $\mathcal{H}(\cdot,\cdot)$ computes the Hamming distance between two binary vectors, and $\gamma$ is a weighting parameter. As shown in Fig. \ref{fig:performance_improvement}, HE \cite{Hamming} and its weighted version \cite{jegou2010improving} improves accuracy considerably in 2008 and 2010.

Applications of HE include video copy detection \cite{douze2010image}, image classification \cite{jain2012hamming} and re-ranking \cite{tolias2014visual}. For example, in image classification, patch matching similarity is efficiently estimated by HE which is integrated into linear kernel-based SVM \cite{jain2012hamming}. In image re-ranking, Tolias \emph{et al.} \cite{tolias2014visual}  use lower HE thresholds to find strict correspondences which resemble those found by RANSAC, and the resulting image subset is more likely to contain true positives for query reformulation.


The improvement over HE has been observed in a number of works, especially from the view of match kernel \cite{selective_match}. To reduce the information loss on the query side, 
Jain \emph{et al.} \cite{jain2011asymmetric} propose a vector-to-binary distance comparison. It exploits the vector-to-hyperplane distance while retaining the efficiency of the inverted index. Further, Qin \emph{et al.} \cite{qin2013query} design a higher-order match kernel within a probabilistic framework and adaptively normalize the local feature distances by the distance distribution of false matches. This method is in the spirit similar to \cite{jegou2010accurate}, in which the word-word distance, instead of the feature-feature distance \cite{qin2013query}, is normalized, according to the neighborhood distribution of each visual word. While the average distance between a word to its neighbors is regularized to be almost constant in \cite{jegou2010accurate}, the idea of democratizing the contribution of individual embeddings has later been employed in \cite{jegou2014triangulation}. In \cite{selective_match}, Tolias \emph{et al.} show that VLAD and HE share similar natures and  propose a new match kernel which trades off between local feature aggregation and feature-to-feature matching, using a similar matching function to \cite{qin2013query}. They also demonstrate that using more bits (\emph{e.g.,} 128) in HE is superior to the original 64 bits scheme at the cost of decreased efficiency. Even more bits (256) are used in \cite{zhou2012scalar}, but this method may be prone to relatively low recall.



\begin{figure}[t]
  \centering
  \includegraphics[width=3.4in]{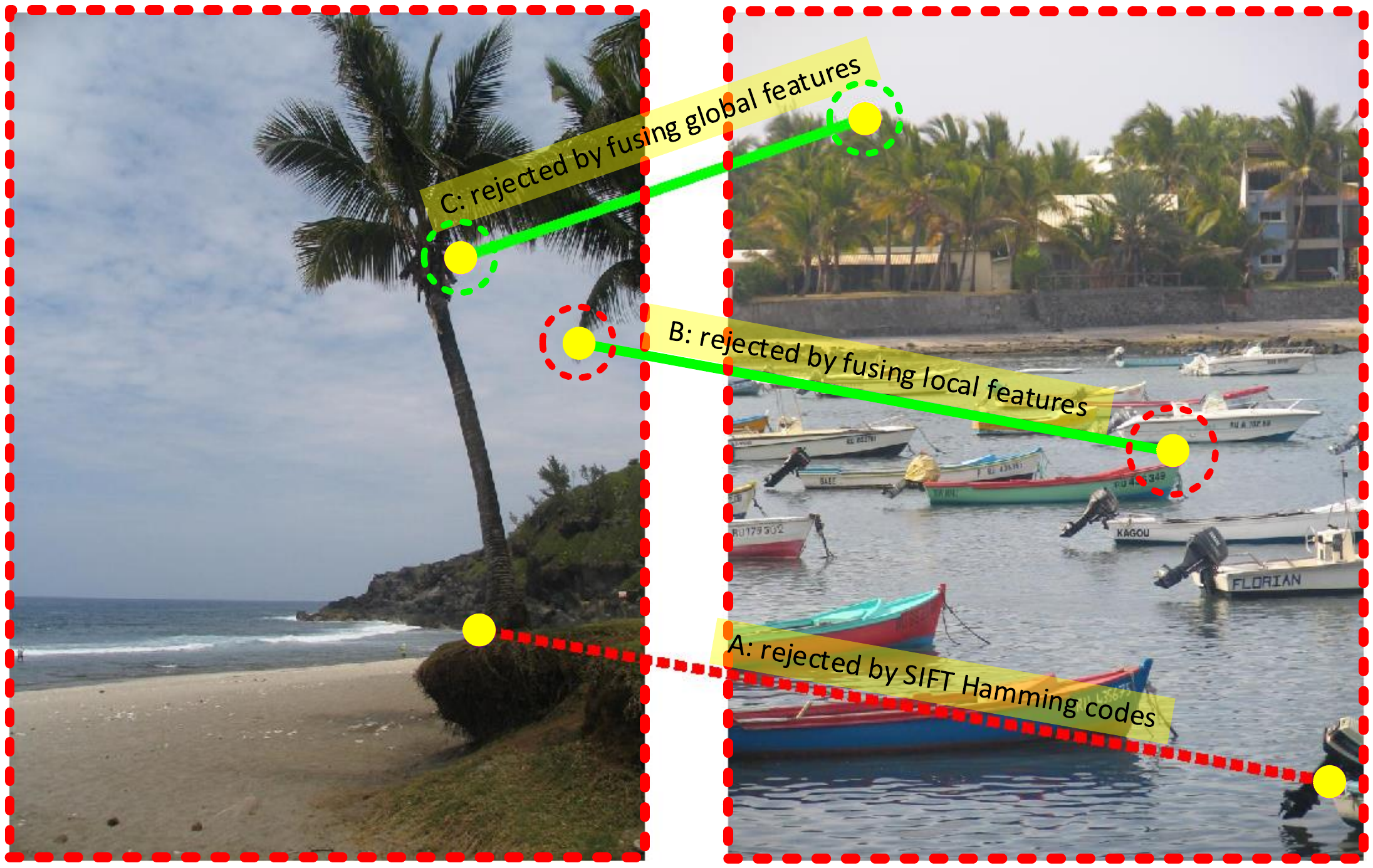}\\
  \caption{False match removal by (A) HE \cite{Hamming}, (B) local-local feature fusion, and (C) local-global feature fusion. }\label{fig:false_match}
\end{figure}

\subsection{Other Important Issues}\label{sec:issues}
\subsubsection{Feature Fusion}\label{sec:fusion_sift}
\textbf{Local-local fusion. } A problem with the SIFT feature is that only local gradient description is provided. Other discriminative information encoded in an image is still not leveraged.
In Fig. \ref{fig:false_match} (B), a pair of false matches cannot be rejected by HE due to their similarity in the SIFT space, but the fusion of other local (or regional) features may correct this problem. A good choice for local-local fusion is to couple SIFT with color descriptors. The usage of color-SIFT descriptors can partially address the trade-off between invariance and discriminative ability. Evaluation has been conducted on several recognition benchmarks \cite{van2010evaluating} of the descriptors such as HSV-SIFT \cite{bosch2008scene}, HueSIFT \cite{van2006boosting} and OpponentSIFT \cite{van2010evaluating}. 
Both HSV-SIFT and HueSIFT are scale-invariant and shift-invariant. OpponentSIFT describes all the channels in the opponent color space using the SIFT descriptor and is largely robust to the light color changes. In \cite{van2010evaluating}, OpponentSIFT is recommended when no prior knowledge about the datasets is available. In more recent works, the binary color signatures are stored in the inverted index \cite{BOC}, \cite{zheng2014packing}. 
Despite the good retrieval accuracy on some datasets, the potential problem is that intensive variation in illumination may compromise the effectiveness of colors.


\textbf{Local-global fusion.} Local and global features describe images from different aspects and can be complementary. In Fig. \ref{fig:false_match} (C), when local (and regional) cues are not enough to reject a false match pair, it would be effective to further incorporate visual information from a larger context scale.
Early and late fusion are two possible ways. In early fusion, the image neighborhood relationship mined by global features such as FC8 in AlexNet \cite{krizhevsky2012imagenet} is fused in the SIFT-based inverted index \cite{co_indexing}. In late fusion, Zhang \emph{el al.} \cite{zhang2012query} build an offline graph for each type of feature, which is subsequently fused during the online query. In an improvement of \cite{zhang2012query}, Deng \emph{et al.} \cite{dengvisual} add weakly supervised anchors to aid graph fusion. Both works on the rank level.
For score-level fusion, automatically learned category-specific attributes are combined with pre-trained category-level information \cite{tao2015attributes}. Zheng \emph{et al.} \cite{zheng2015query} propose the query-adaptive late fusion by  extracting a number of features (local or global, good or bad) and weighting them in a query-adaptive manner.

\subsubsection{Geometric Matching}
A frequent concern with the BoW model is the lack of geometric constraints among local features. Geometric verification can be used as a critical pre-processing step various scenarios, such as query expansion \cite{chum2007total,chum2011total}, feature selection \cite{turcot2009better}, database-side feature augmentation \cite{turcot2009better}, \cite{root_sift}, large-scale object mining \cite{philbin2011geometric}, \emph{etc}.
The most well-known method for global spatial verification is RANSAC \cite{AKM}. It calculates affine transformations for each correspondence repeatedly which are verified by the number of inliers that fit the transformation. RANSAC is effective in re-ranking a subset of top-ranked images but has efficiency problems. As a result, how to efficiently and accurately incorporate spatial cues in the SIFT-based framework has been extensively studied.

A good choice is to discover the spatial context among local features. For example, visual phrases \cite{zheng2006effective, DVP, sadeghi2011recognition, jiang2012randomized} are generated among individual visual words to provide more strict matching criterion. Visual word co-occurrences in the entire image are estimated \cite{prefilters} and aggregated \cite{koniusz2016higher}, while in \cite{collocation, triplets}, \cite{wu2009bundling} visual word clusters within local neighborhoods are discovered. Visual phrases can also be constructed from adjacent image patches \cite{zheng2006effective}, random spatial partitioning \cite{jiang2012randomized}, and localized stable regions \cite{wu2009bundling} such as MSER \cite{matas2004robust}.

Another strategy uses voting to check geometric consistency. In the voting space, a bin with a larger value is more likely to represent the true transformation. An important work is weak geometrical consistency (WGC) \cite{Hamming}, which focuses on the difference in scale and orientation between matched features. The space of difference is  quantized into bins. Hough voting is used to locate the subset of correspondences similar in scale or orientation differences. Many later works can be viewed as extensions of WGC. For example, the method of Zhang \emph{et al.} \cite{GVP} can be viewed as WGC using x, y offsets instead of scale and orientation. This method is invariant to object translations, but may be sensitive to scale and rotation changes due to the rigid coordinate quantization. To regain the scale and the rotation variance, Shen \emph{et al.} \cite{shen2012object} quantize the angle and scale of the query region after applying several transformations. A drawback of \cite{shen2012object} is that query time and memory cost are both increased. To enable efficient voting and alleviate quantization artifacts, Hough pyramid matching (HPM) \cite{avrithis2014hough} distributes the matches over a hierarchical partition of the transformation space. HPM trades off between flexibility and accuracy and is very efficient.  Quantization artifact can also be reduced by allowing a single correspondence to vote for multiple bins \cite{wu2015adaptive}. HPM and \cite{wu2015adaptive} are much faster than RANSAC and can be viewed as extensions in the rotation and the scale invariance to the weak geometry consistency proposed along with Hamming Embedding \cite{Hamming}. In \cite{li2015pairwise}, a rough global estimate of orientation and scale changes is made by voting, which is used to verify the transformation obtained by the matched features. A recent method \cite{schonberger2016vote} combines the advantage of hypothesis-based methods such as RANSAC \cite{AKM} and voting-based methods \cite{avrithis2014hough,GVP,wu2015adaptive,li2015pairwise}. Possible hypothesises are identified by voting and later verified and refined. This method inherits efficiency from voting and supports query expansion since it outputs an explicit transformation and a set of inliers. 

\subsubsection{Query Expansion}
As a post-processing step, query expansion (QE) significantly improves the retrieval accuracy. In a nutshell, a number of top-ranked images from the original rank list are employed to issue a new query which is in turn used to obtain a new rank list. QE allows additional discriminative features to be added to the original query, thus improving recall. 

In instance retrieval, Chum \emph{et al.} \cite{chum2007total} are the first to exploit this idea. They propose the average query expansion (AQE) which averages features of the top-ranked images to issue the new query. Usually, spatial verification \cite{AKM} is employed for re-ranking and obtaining the ROIs from which the local features undergo average pooling. AQE is used by many later works \cite{gordo2016deep,tolias2016particular,radenovic2016cnn}  as a standard tool. The recursive AQE and the scale-band recursive QE are effective improvement but incur more computational cost \cite{chum2007total}. Four years later, Chum \emph{et al.} \cite{chum2011total} improve QE from the perspectives of learning background confusers, expanding the query region and incremental spatial verification. In \cite{root_sift}, a linear SVM is trained online using the top-ranked and bottom-ranked images as positive and negative training samples, respectively. The learned weight vector is used to compute the average query. Other important extensions include ``hello
neighbor'' based on reciprocal neighbors \cite{qin2011hello}, QE with rank-based weighting \cite{shen2012object}, Hamming QE \cite{tolias2014visual} (see Section \ref{sec:medium_codebook}), \emph{etc.} 

\subsubsection{Small Object Retrieval}
Retrieving objects that cover a small portion of images is a challenging task due to 1) the few detected local features and 2) the large amount of background noise. The Instance Search task in the TRECVID campaign \cite{awad2016trecvid} and the task of logo retrieval are important venues/applications for this task. 

Generally speaking, both TRECVID and logo retrieval can be tackled with similar pipelines. For keypoint-based methods, the spatial context among the local features is important to discriminative target objects from others, especially in cases of rigid objects. Examples include  \cite{chum2009geometric,kalantidis2011scalable,romberg2013bundle}. Other effective methods include burstiness handling \cite{revaud2012correlation} (discussed in Section \ref{sec:tf-idf}), considering the different inlier ratios between the query and target objects \cite{zhu2013query}, \emph{etc.} In the second type of methods,  effective region proposals \cite{chen2017retrieving} or multi-scale image patches \cite{iscen2017efficient} can be used as object region candidates. In \cite{iscen2017efficient}, a recent state-of-the-art method, a regional diffusion mechanism based on neighborhood graphs is proposed to further improve the recall of small objects. 




\section{CNN-based Image Retrieval}\label{sec:cnn_model}
CNN-based retrieval methods have constantly been  proposed in recent years and are gradually replacing the hand-crafted local detectors and descriptors. In this survey, CNN-based methods are classified into three categories: using pre-trained CNN models, using fine-tuned CNN models and hybrid methods. The first two categories compute the global feature with a single network pass, and the hybrid methods may require multiple network passes (see Fig. \ref{fig:pipeline}).
\subsection{Retrieval Using Pre-trained CNN Models}
This type of methods is efficient in feature computation due to the single-pass mode. Given the transfer nature, its success lies in the feature extraction and encoding steps. We will first describe some commonly used datasets and networks for pre-training, and then the feature computation process.
\subsubsection{Pre-trained CNN Models}\label{sec:pretrained_cnn}
\textbf{Popular CNN architectures.} Several CNN models serve as good choices for extracting features, including AlexNet \cite{krizhevsky2012imagenet}, VGGNet \cite{simonyan2014very}, GoogleNet \cite{szegedy2015going} and ResNet \cite{he2016deep}, which are listed in Table \ref{table:CNN-models}. Briefly, CNN can be viewed as a set of non-linear functions and is composed of a number of layers such as convolution, pooling, non-linearities, \emph{etc}. CNN has a hierarchical structure. From bottom to top layers, the image undergoes convolution with filters, and the receptive field of these image filters increases. Filters in the same layer have the same size but different parameters. AlexNet \cite{krizhevsky2012imagenet} was proposed the earliest among these networks, which has five convolutional layers and three fully connected (FC) layers. It has 96 filters in the first layer of sizes $11\times11\times3$ and has 256 filters of size $3\times 3\times 192$ in the 5th layer. Zeiler \emph{et al.} \cite{zeiler2014visualizing} observe that the filters are sensitive to certain visual patterns and that these patterns evolve from low-level bars in bottom layers to high-level objects in top layers. For low-level and simple visual stimulus, the CNN filters act as the detectors in the local hand-crafted features, but for the high-level and complex stimulus, the CNN filters have distinct characteristics that depart from SIFT-like detectors. AlexNet has been shown to be outperformed by newer ones such as VGGNet, which has the largest number of parameters. ResNet and GoogleNet won the ILSVRC 2014 and 2015 challenges, respectively, showing that CNNs are more effective with more layers.  A full review of these networks is beyond the scope of this paper, and we refer readers to \cite{krizhevsky2012imagenet,chatfield2014return}, \cite{simonyan2014very} for details.

\textbf{Datasets for pre-training. }Several large-scale recognition datasets are used for CNN pre-training. Among them, the ImageNet dataset \cite{deng2009imagenet} is mostly commonly used. It contains 1.2 million images of 1000 semantic classes and is usually thought of as being generic. Another data source for pre-training is the Places-205 dataset \cite{zhou2014learning} which is twice as large as ImageNet but has five times fewer classes. It is a scene-centric dataset depicting various indoor and outdoor scenes. A hybrid dataset combining the Places-205 and the ImageNet datasets has also been used for pre-training \cite{zhou2014learning}. The resulting HybridNet is evaluated in \cite{azizpour2016factors,chandrasekhar2016practical,szegedy2015going,he2016deep} for instance retrieval.

\setlength{\tabcolsep}{3.45pt}
\begin{table}[t]
\footnotesize
\caption{Pre-trained CNN models that can be used.}
\begin{tabular}{l|lllll}
\hline
\multicolumn{1}{l|}{models} &size & \# layers & training Set &used in\\
\hline
OverFeat \cite{sermanet2014overfeat} & \multirow{1}{*}{144M}& \multirow{1}{*}{6+3} & ImageNet& \cite{razavian2014cnn} \\
\hline
AlexNet \cite{krizhevsky2012imagenet} & \multirow{3}{*}{60M}& \multirow{3}{*}{5+3} & ImageNet& \cite{gong2014multi,razavian2016paper} \\
PlacesNet \cite{zhou2014learning}&&&Places& \cite{azizpour2016factors,chandrasekhar2016practical}\\
HybridNet \cite{zhou2014learning}&&&ImageNet+Places& \cite{azizpour2016factors,chandrasekhar2016practical}\\
\hline
\multicolumn{1}{l|}{VGGNet \cite{simonyan2014very}} &138M & 13+3 &ImageNet&\cite{tolias2016particular} \\
\hline
\multicolumn{1}{l|}{GoogleNet \cite{szegedy2015going}} &11M& 22 & ImageNet&\cite{ng2015exploiting}\\
\hline
\multicolumn{1}{l|}{ResNet \cite{he2016deep}} &44.6M& 101 & ImageNet& n.a\\

\hline
\end{tabular}
\label{table:CNN-models}
\end{table}

\textbf{The transfer issue.} Comprehensive evaluation of various CNNs on instance retrieval has been conducted in several recent works \cite{azizpour2016factors,chandrasekhar2016practical,zheng2016good}. The transfer effect is mostly concerned. It is considered in \cite{azizpour2016factors} that instance retrieval, as a target task, lies farthest from the source, \emph{i.e.,} ImageNet. Studies reveal some critical insights in the transfer process. First, during model transfer, features extracted from different layers exhibit different retrieval performance. Experiments confirm that the top layers may exhibit lower generalization ability than the layer before it. For example, for AlexNet pre-trained on ImageNet, it is shown that FC6, FC7, and FC8 are in descending order regarding retrieval accuracy \cite{azizpour2016factors}. It is also shown in \cite{zheng2016good,tolias2016particular} that the pool5 feature of  AlexNet and VGGNet is even superior to FC6 when proper encoding techniques are employed. Second, the source training set is relevant to retrieval accuracy on different datasets. For example, Azizpour \emph{et al.} \cite{azizpour2016factors} report that HybridNet yields the best performance on Holidays after PCA. They also observe that AlexNet pre-trained on ImageNet is superior to PlacesNet and HybridNet on the Ukbench dataset \cite{HKM} which contains common objects instead of architectures or scenes.  So the similarity of the source and target plays a critical role in instance retrieval when using a pre-trained CNN model.

\subsubsection{Feature Extraction}

\textbf{FC descriptors. }The most straightforward idea is to extract the descriptor from the fully-connected (FC) layer of the network \cite{razavian2014cnn,babenko2014neural,zheng2016accurate}, \emph{e.g.,} the 4,096-dim FC6 or FC7 descriptor in AlexNet. The FC descriptor is generated after layers of convolutions with the input image, has a global receptive field, and thus can be viewed as a global feature. It yields fair retrieval accuracy under Euclidean distance and can be improved with power normalization \cite{perronnin2010improving}.

\textbf{Intermediate local features. } Many recent retrieval methods \cite{tolias2016particular,ng2015exploiting,zheng2016good}  focus on local descriptors in the intermediate layers. In these methods, lower-level convolutional filters (kernels) are used to detect local visual patterns. Viewed as local detectors, these filters have a smaller receptive field and are densely applied on the entire image. Compared with the global FC feature, local detectors are more robust to image transformations such as truncation and occlusion, in ways that are similar to the local invariant detectors (Section \ref{sec:local_feature_extraction}).

Local descriptors are tightly coupled with these intermediate local detectors, \emph{i.e.,} they are the responses of the input image to these convolution operations.  In other words, after the convolutions, the resulting activation maps can be viewed as a feature ensemble, which is called the ``column feature'' in this survey. For example in AlexNet \cite{krizhevsky2012imagenet}, there are $n=96$ detectors (convolutional filters) in the 1st convolutional layer. These filters produces $n=96$ heat maps of size $27\times27$ (after max pooling). Each pixel in the maps has a receptive field of $19\times19$ and records the response of the image \emph{w.r.t} the corresponding filter \cite{tolias2016particular,ng2015exploiting,zheng2016good}. The column feature is therefore of size $1\times1\times96$ (Fig. \ref{fig:pipeline}) and can be viewed as a description of a certain patch in the original image.  Each dimension of this descriptor denotes the level of activation of the corresponding detector and resembles the SIFT descriptor to some extent.  The column feature initially appears in \cite{razavian2016paper}, where Razavian \emph{et al.} first do max-pooling over regularly partitioned windows on the feature maps and then concatenate them across all filter responses, yielding column-like features. In \cite{hariharan2015hypercolumns}, column features from multiple layers of the networks are concatenated, forming the ``hypercolumn'' feature.

\subsubsection{Feature Encoding and Pooling}
When column features are extracted, an image is represented by a set of descriptors. To aggregate these descriptors into a global representation, currently two strategies are adopted: encoding and direct pooling (Fig. \ref{fig:pipeline}).

\emph{Encoding.} A set of column features resembles a set of SIFT features. So standard encoding schemes can be directly employed. The most commonly used methods are VLAD \cite{jegou2010aggregating} and FV \cite{perronnin2010improving}. A brief review of VLAD and FV can be seen in Section \ref{sec:encoding_small}. A milestone work is \cite{ng2015exploiting}, in which the column features are encoded into VLAD for the first time. This idea was later extended to CNN model fine-tuning \cite{arandjelovic2016netvlad}. The BoW encoding can also be leveraged, as the case in \cite{kulkarni2015hybrid}. The column features within each layer are aggregated into a BoW vector which is then concatenated across the layers.  An exception to these fix-length representations is \cite{mohedano2016bags}, in which the column features are quantized with a codebook of size 25k and an inverted index is employed for efficiency.


\emph{Pooling.} A major difference between the CNN column feature and SIFT is that the former has an explicit meaning in each dimension, \emph{i.e.,} the response of a particular region of the input image to a filter. Therefore, apart from the encoding schemes mentioned above, direct pooling techniques can produce discriminative features as well.

A milestone work in this direction consists in the Maximum activations of convolutions (MAC) proposed by Tolias \emph{et al.} \cite{tolias2016particular}. Without distorting or cropping images, MAC computes a global descriptor with a single forward pass. Specifically, MAC calculates the maximum value of each intermediate feature map and concatenates all these values within a convolutional layer. In its multi-region version, the integral image and an approximate maximum operator are used for fast computation. The regional MAC descriptors are subsequently sum-pooled along with a series of normalization and PCA-whitening operations \cite{jegou2012negative}. 
We also note in this survey that several other works \cite{babenko2015aggregating}, \cite{razavian2016paper,zheng2016good} also employ similar ideas with \cite{tolias2016particular} in employing max or average pooling on the intermediate feature maps and that Razavian \emph{et al.} \cite{razavian2016paper} are the first. It has been observed that the last convolutional layer (\emph{e.g.,} pool5 in VGGNet), after pooling usually yields superior accuracy to the FC descriptors and the other convolutional layers \cite{zheng2016good}.

Apart from direct feature pooling,
it is also beneficial to assign some specific weights to the feature maps within each layer before pooling. In \cite{babenko2015aggregating}, Babenko \emph{et al.} propose the injection of the prior knowledge that objects tend to be located toward image centers, and impose a 2-D Gaussian mask on the feature maps before sum pooling. Xie \emph{et al.} \cite{xie2016interactive} improve the MAC representation \cite{tolias2016particular} by propagating the high-level semantics and spatial context to low-level neurons for improving the descriptive ability of these bottom-layer activations. With a more general weighting strategy, Kalantidis \emph{et al.} \cite{kalantidis2016cross} perform both feature map-wise and channel-wise weighing, which aims to highlight the highly active spatial responses while reducing burstiness effects.

\subsection{ Image Retrieval with Fine-Tuned CNN Models}\label{sec:ft_cnn_model}
Although pre-trained CNN models have achieved impressive retrieval performance, a hot topic consists in fine-tuning the CNN model on specific training sets. When a fine-tuned CNN model is employed, the image-level descriptor is usually generated in an end-to-end manner, \emph{i.e.,} the network will produce a final visual representation without additional explicit encoding or pooling steps.

\subsubsection{Datasets for Fine-Tuning}
The nature of the datasets used in fine-tuning is the key to learning discriminative CNN features. ImageNet \cite{deng2009imagenet} only provides images with class labels. So the pre-trained CNN model is competent in discriminating images of different object/scene classes, but may be less effective to tell the difference between images that fall in the same class (\emph{e.g.,} architecture)  but depict different instances (\emph{e.g.,} ``Eiffel Tower'' and ``Notre-Dame''). Therefore, it is important to fine-tune the CNN model on task-oriented datasets.

\setlength{\tabcolsep}{3.55pt}
\begin{table}[t]
\centering
\caption{Statistics of instance-level datasets having been used in fine-tuning.}
\begin{tabular}{l|ccl}
\hline
name&  \# images & \# classes & content \\
\hline
\hline
Landmarks \cite{babenko2014neural}& 213,678 & 672 &Landmark \\
3D Landmark \cite{radenovic2016cnn}& 163,671 & 713 & Landmark \\
Tokyo TM \cite{arandjelovic2016netvlad}& 112,623& n.a & Landmark \\
MV RGB-D \cite{lai2011large}& 250,000 & 300 &House. object \\
Product \cite{bell2015learning}& 101,945$\times$2& n.a &Furniture\\
\hline
\end{tabular}
\label{table:ft_datasets}
\end{table}

The datasets having been used for fine-tuning in recent years are shown in Table \ref{table:ft_datasets}. Buildings and common objects are the focus. The milestone work on fine-tuning is \cite{babenko2014neural}. It collects the \textbf{Landmarks dataset} by a semi-automated approach: automated searching for the popular landmarks in Yandex search engine, followed by a manual estimation of the proportion of relevant image among the top ranks. This dataset contains 672 classes of various architectures, and  the fine-tuned network produces superior features on landmark related datasets such as Oxford5k \cite{AKM} and Holidays \cite{Hamming}, but has decreased performance on  Ukbench \cite{HKM} where common objects are presented. Babenko \emph{et al.} \cite{babenko2014neural} have also fine-tuned CNNs on the \textbf{Multi-view RGB-D dataset} \cite{lai2011large} containing turntable views of 300 household objects, in order to improve performance on Ukbench. The Landmark dataset is later used by Gordo \emph{et al.} \cite{gordo2016deep} for fine-tuning, after an automatic cleaning approach based on SIFT matching.
In \cite{radenovic2016cnn}, Radenovi{\'c} \emph{et al.} employ the retrieval and Structure-From-Motion methods to build \textbf{3D landmark} models so that images depicting the same architecture can be grouped. Using this labeled dataset, the linear discriminative projections (denoted as L$_w$ in Table \ref{table:compare_methods_datasets}) outperform the previous whitening technique \cite{jegou2012negative}. Another dataset called \textbf{Tokyo Time Machine} is collected using Google Street View Time Machine which provides images depicting the same places over time \cite{arandjelovic2016netvlad}.  While most of the above datasets focus on landmarks, Bell \emph{et al.} \cite{bell2015learning} build a \textbf{Product dataset} consisting of furniture by developing a crowd-sourced pipeline to draw connections between in-situ objects and the corresponding products. It is also feasible to fine-tune on the query sets suggested in \cite{salvador2016faster}, but this method may not be adaptable to new query types.

\subsubsection{Networks in Fine-Tuning}
The CNN architectures used in fine-tuning mainly fall into two types: the classification-based network and the verification-based network. The classification-based network is trained to classify architectures into pre-defined categories. Since there is usually no class overlap between the training set and the query images, the learned embedding \emph{e.g.,} FC6 or FC7 in AlexNet, is used for Euclidean distance based retrieval. This train/test strategy is employed in \cite{babenko2014neural}, in which the last FC layer is modified to have 672 nodes corresponding to the number of classes in the Landmark dataset.

The verification network may either use  a siamese network with pairwise loss or use a triplet loss and has been more widely employed for fine-tuning. A standard siamese network based on AlexNet and the contrastive loss is employed in \cite{bell2015learning}.  In \cite{radenovic2016cnn}, Radenovi{\'c} \emph{et al.}  propose to replace the FC layers with a MAC layer \cite{tolias2016particular}. Moreover, with the 3D architecture models built in \cite{radenovic2016cnn}, training pairs can be mined. Positive image pairs are selected based on the number of co-observed 3D points (matched SIFT features), while hard negatives are defined as those with small distances in their CNN descriptors. These image pairs are fed into the siamese network, and the contrastive loss is calculated from the $\ell_2$ normalized MAC features. In a concurrent work to \cite{radenovic2016cnn}, Gordo \emph{et al.} \cite{gordo2016deep} fine-tune a triplet-loss network and a region proposal network on the Landmark dataset \cite{babenko2014neural}. The superiority of \cite{gordo2016deep} consists in its localization ability, which excludes the background in feature learning and extraction. In both works, the fine-tuned models exhibit state-of-the-art accuracy on landmark retrieval datasets including Oxford5k, Paris6k and Holidays, and also good generalization ability on Ukbench (Table \ref{table:compare_methods_datasets}). In \cite{arandjelovic2016netvlad}, a VLAD-like layer is plugged in the network at the last convolutional layer which is amenable to training via back-propagation. Meanwhile, a new triplet loss is designed to make use of the weakly supervised Google Street View Time Machine data.

\subsection{Hybrid CNN-based Methods}\label{sec:hybrid}
For the hybrid methods, multiple network passes are performed. A number of image patches are generated from an input image, which are fed into the network for feature extraction before an encoding/pooling stage. Since the manner of ``detector + descriptor'' is similar to SIFT-based methods, we call this method type ``hybrid''. It is usually less efficient than the single-pass methods.
\subsubsection{Feature Extraction}
In hybrid methods, the feature extraction process consists of patch detection and description steps. For the first step, the literature has seen three major types of region detectors. The first is grid image patches. For example, in \cite{gong2014multi}, a two-scale sliding window strategy is employed to generate patches. In \cite{razavian2014cnn}, the dataset images are first cropped and rotated, and then divided into patches of different scales, the union of which covers the whole image. The second type is invariant keypoint/region detectors. For instance, the difference of Gaussian feature points are used in \cite{zagoruyko2015learning}; the MSER region detector is leveraged in \cite{fischer2014descriptor}. Third, region proposals also provide useful information on the locations of the potential objects. Mopuri \emph{et al.} \cite{mopuri2015object} employ selective search \cite{uijlings2013selective} to generate image patches, while EdgeBox \cite{zitnick2014edge} is used in  \cite{uricchio2015fisher}. In \cite{salvador2016faster}, the region proposal network (RPN) \cite{ren2015faster} is applied to locate the potential objects in an image.


The use of CNN as region descriptors is validated in \cite{fischer2014descriptor}, showing that CNN is superior to SIFT in image matching except on blurred images.
Given the image patches, the hybrid CNN method usually employs the FC or pooled intermediate CNN features. Examples using the FC descriptors include \cite{gong2014multi,razavian2014cnn,xie2015image,mopuri2015object}. In these works, the 4,096-dim FC features are extracted from the multi-scale image regions \cite{gong2014multi,razavian2014cnn,xie2015image} or object proposals \cite{mopuri2015object}.
On the other hand, Razavian \emph{et al.} \cite{razavian2016paper} also uses the intermediate descriptors after max-pooling as region descriptors. 

The above methods use pre-trained models for patch feature extraction. Based on the hand-crafted detectors, patch descriptors can also be learned through CNN in either supervised \cite{paulin2015local} or unsupervised manner \cite{zagoruyko2015learning}, which improves over the previous works on SIFT descriptor learning \cite{philbin2010descriptor}, \cite{simonyan2014learning}. Yi \emph{et al.} \cite{yi2016lift} further propose an end-to-end learning method integrating region detector, orientation estimator and feature descriptor in a single pipeline.

\subsubsection{Feature Encoding and Indexing}
The encoding/indexing procedure of hybrid methods resembles  SIFT-based retrieval, \emph{e.g., } VLAD/FV encoding under a small codebook or the inverted index under a large codebook.

The VLAD/FV encoding, such as \cite{gong2014multi,mopuri2015object}, follow the standard practice in the case of SIFT features \cite{jegou2010aggregating,perronnin2010improving}, so we do not detail here.
On the other hand, several works exploit the inverted index on the patch-based CNN features \cite{liu2015deepindex,li2016exploiting,mohedano2016bags}. Again, standard techniques in SIFT-based methods such as HE are employed \cite{li2016exploiting}.
Apart from the above-mentioned strategies, we notice that several works \cite{razavian2014cnn,razavian2016paper,xie2015image} extract several region descriptors per image to do a many-to-many matching, called ``spatial search'' \cite{razavian2014cnn}. This method improves the translation and scale invariance of the retrieval system but may encounter efficiency problems. A reverse strategy to applying encoding on top of CNN activations is to build a CNN structure (mainly consisting of FC layers) on top of SIFT-based representations such as FV. By training a classification model on natural images, the intermediate FC layer can be used for retrieval \cite{perronnin2015fisher}.

\subsection{Discussions}
\subsubsection{Relationship between SIFT- and CNN-based Methods}
In this survey, we categorize current literature into six fine-grained classes.
The differences and some representative works of the six categories are summarized in Table \ref{table:three_types_methods} and Table \ref{table:compare_methods_datasets}. Our observation goes below.

First, the hybrid method can be viewed as a transition zone from SIFT- to CNN-based methods. It resembles the SIFT-based methods in all the aspects except that it extracts CNN features as the local descriptor. Since the network is accessed multiple times during patch feature extraction, the efficiency of the feature extraction step may be compromised.

Second, the single-pass CNN methods tend to combine the individual steps in the SIFT-based and hybrid methods. In Table \ref{table:compare_methods_datasets}, the ``pre-trained single-pass'' category integrates the feature detection and description steps; in the ``fine-tuned single-pass'' methods, the image-level descriptor is usually extracted in an end-to-end mode, so that no separate encoding process is needed. In \cite{gordo2016deep}, a ``PCA'' layer is integrated for discriminative dimension reduction, making a further step towards end-to-end feature learning.

Third, fixed-length representations are gaining more popularity due to efficiency considerations. It can be obtained by aggregating local descriptors (SIFT or CNN) \cite{jegou2010aggregating}, \cite{jegou2014triangulation,gong2014multi}, \cite{ng2015exploiting}, direct pooling \cite{mopuri2015object}, \cite{tolias2016particular}, or end-to-end feature computation \cite{babenko2014neural,gordo2016deep}. Usually, dimension reduction methods such as PCA can employed on top of the fixed-length representations, and ANN search methods such as PQ \cite{jegou2010aggregating} or hashing \cite{perronnin2010large} can be used for fast retrieval.

\subsubsection{Hashing and Instance Retrieval}
Hashing is a major solution to the approximate nearest neighbor problem. It can be categorized into locality sensitive hashing (LSH) \cite{indyk1998approximate} and learning to hash. LSH is data-independent and is usually outperformed by learning to hash, a data-dependent hashing approach. For learning to hash, a recent survey \cite{wang2016survey} categorizes it into quantization and pairwise similarity preserving. The quantization methods are briefly discussed in Section \ref{sec:encoding_small}. For the pairwise similarity preserving methods, some popular hand-crafted methods include Spectral hashing \cite{weiss2009spectral}, LDA hashing \cite{strecha2012ldahash}, \emph{etc}.

Recently, hashing has seen a major shift from hand-crafted to supervised hashing with deep neural networks. These methods take the original image as input and produce a learned feature before binarization \cite{lin2015deep,zhao2015deep}. Most of these methods, however, focus on class-level image retrieval, a different task with instance retrieval discussed in this survey. For instance retrieval, when adequate training data can be collected, such as architecture and pedestrians, the deep hashing methods may be of critical importance.

\setlength{\tabcolsep}{4.55pt}
\begin{table}[t]
\centering
\caption{Statistics of popular instance-level datasets.}
\begin{tabular}{l|ccl}
\hline
name&  \# images & \# queries & content \\
\hline
\hline
Holidays \cite{Hamming}& 1,491 & 500 &scene \\
Ukbench \cite{HKM}& 10,200 & 10,200 & common objects \\
Paris6k \cite{soft}& 6,412& 55 & buildings \\
Oxford5k \cite{AKM}& 5,062&55 &buildings\\
\multirow{2}{*}{Flickr100k \cite{soft}} &\multirow{2}{*}{99,782}&\multirow{2}{*}{-}&\multirow{2}{0.8in}{from Flickr's popular tags}\\
&&&\\
\hline
\end{tabular}
\label{table:compare_datasets}
\end{table}
\section{Experimental Comparisons}\label{sec:experiment}
\subsection{Image Retrieval Datasets}
Five popular instance retrieval datasets are used in this survey. Statistics of these datasets can be accessed in Table \ref{table:compare_datasets}.

\setlength{\tabcolsep}{1.44pt}
\begin{sidewaystable*}[htbp]
\renewcommand{\arraystretch}{1.3}
\caption{Performance summarization of some representative methods of the six categories on the benchmarks. ``+100k'' --> the addition of Flickr100k into Oxford5k. ``pw.'' --> power low normalization \cite{perronnin2010improving}. ``MP'' --> max pooling.  ``SP'' --> sum pooling. * and parentheses --> results are obtained with post-processing steps such as spatial verification or QE.  $^{\S}$ --> numbers are estimated from the curves. $^{\dag}$ numbers are reported by our implementation. For Holidays, results using the rotated images are presented after ``/''. For Oxford5k (+100k) and Paris6k, results using the full-sized queries are shown after ``/''. $^{\natural}$ --> the full query image is fed into the network, but only the features whose centers fall into the query region of interest are aggregated. Note that in many fixed-length representations, ANN algorithms such as PQ are not used to report the results, but ANN can be readily applied after PCA during indexing. }
\small
\centering
\begin{tabular}{c|c|l|l|l|l|l|c|c|cccccc}
\hline
\multicolumn{2}{l|}{method type} &  methods& detector & descriptor &encoding &indexing&voc&dim & Holidays & Ukbench & Oxford5k&+100k&Paris6k & mem/img \\
\hline
\hline
\multirow{15}{*}{\rotatebox{90}{SIFT-based}} & \multirow{5}{*}{{Large Voc.}} & HKM \cite{HKM}&MSER&SIFT&hier. soft, BoW& inv. index & \multicolumn{2}{c|}{1M} &59.7 &2.85& 44.3$^{\dag}$ &26.6$^{\dag}$&46.5$^{\dag}$& 9.8kb\\
 && \multirow{1}{*}{AKM \cite{AKM}} &hes-aff&SIFT&hard, BoW& inv. index& \multicolumn{2}{c|}{1M}& 64.1$^{\dag}$ & 3.02$^{\dag}$ &49.3&34.3&50.2$^{\dag}$ &9.8kb \\
&& Fine Voc. \cite{mikulik2010learning}& hes-aff&SIFT & alt. word, BoW & inv. index & \multicolumn{2}{c|}{16M} &-/74.9 (75.8) & - &74.2 (84.9)&67.4 (79.5)&74.9 (82.4)&9.8kb\\
&& Three Things \cite{root_sift}& hes-aff &rootSIFT&hard, BoW& inv. index & \multicolumn{2}{c|}{1M} &-& - & 80.9*&72.2*&76.5*&22.0kb  \\
&& Co-index \cite{co_indexing}  & DoG & SIFT, CNN & hard, BoW & co-index & \multicolumn{2}{c|}{1M}&80.86 &3.60& 68.72 &-&-& 21.6kb  \\
\cline{2-15}
&\multirow{5}{*}{{Mid Voc.}}& \footnotesize HE+WGC \cite{Hamming,jegou2010improving} &\multirow{1}{*}{hes-aff}&\multirow{1}{*}{SIFT}&\multirow{1}{*}{hard, BoW, HE}&\multirow{1}{*}{inv. index}&\multicolumn{2}{c|}{20k} &81.3 (84.2)& 3.42 (3.55) & 61.5 (74.7) &51.6 (68.7)& - &36.8kb  \\
&& \multirow{1}{*}{Burst \cite{burstiness}}&hes-aff&SIFT& burst, BoW, HE & inv. index & \multicolumn{2}{c|}{20k} &83.9 (84.8)& 3.54 (3.64) & 64.7 (68.5)&58$^{\S}$ (63$^{\S}$)&62.8$^{\dag}$ &36.8kb  \\
&& Q.ada \cite{qin2013query}& hes-aff &rootSIFT& MA, BoW, HE & inv. index &  \multicolumn{2}{c|}{20k}&78.0 &-& 82.1&72.8& 73.6& 36.8kb \\
&& ASMK \cite{selective_match} &hes-aff&rootSIFT&MA, BoW, agg.HE& inv. index& \multicolumn{2}{c|}{65k} &81.0 & - & 80.4&75.0 (85.0)&77.0&43.2kb \\
&& c-MI \cite{zheng2014packing} &hes-aff&\footnotesize rootSIFT, HS&MA, BoW, HE& 2D index  & \multicolumn{2}{c|}{20k$\times$200} & 84.0 & 3.71 & 58.2$^{\dag}$&35.2$^{\dag}$&55.1$^{\dag}$&45.2kb \\
\cline{2-15}
&\multirow{5}{*}{{Small Voc.}}& VLAD \cite{jegou2010aggregating}  &hes-aff&SIFT&VLAD&PCA, PQ & 64&4,096 & 55.6& 3.18$^{\dag}$&  37.8&27.2$^{\dag}$&38.6$^{\dag}$&16kb \\

&& FV \cite{perronnin2010large,jegou2012aggregating}  &hes-aff&PCA-SIFT&FV, pw.&LSH, SH & 64&4,096 &59.5& 3.35 & 41.8 &33.1$^{\dag}$&43.0&16kb \\
&& All A. VLAD \cite{arandjelovic2013all}  &hes-aff&SIFT&Improved VLAD&PCA & 64&128 & 62.5& -&  44.8&-&-&0.5kb \\
&& NE \cite{jegou2012negative}&hes-aff& rootSIFT &\footnotesize NE, multi.voc, VLAD& PCA$_w$  & 4$\times$256&128  &61.4&3.36  &  - &-&-&0.5kb\\
&& Triangulation \cite{jegou2014triangulation}&hes-aff& rootSIFT &triang+democ& PCA, pw.  & 16&128  &61.7&3.40  &  43.3 &35.3&-&0.5kb\\
\hline
\hline
 \multirow{14}{*}{\rotatebox{90}{CNN-based}}&\multirow{5}{*}{Hybrid }& Off the Shelf \cite{razavian2014cnn} &\footnotesize den. patch&\footnotesize OFeat 1st FC& - & - & -&-&84.3 & 3.64 & -/68.0 &-&-/79.5& 4-15kb \\
&& MSS \cite{razavian2016paper} &\footnotesize den. patch&vgg conv5& MP & - & - & -&88.1 & 3.72 & -/84.4&-&-/85.3 &16kb\\
&& CKN \cite{paulin2015local} &hes-aff& CKN &VLAD, power&- & 256&65k & 79.3 & 3.76 & 56.5 &-&-&256kb \\
&& MOP \cite{gong2014multi} &\footnotesize den. patch&alex FC7&multi-scale VLAD&PCA$_w$ & 100&2,048&80.2 & - & -&-& -& 8kb  \\
&& OLDFP\cite{mopuri2015object} &\footnotesize sel. search&alex FC7&MP& ITQ \cite{gong2011iterative}& -&512 & 88.5 & 3.81 & 60.7 &-&66.2 &2kb \\
\cline{2-15}
 &\multirow{5}{*}{\tabincell{c}{Pre-trained\\ single-pass} }& BLCF \cite{mohedano2016bags} &\multicolumn{2}{l|}{vgg16, conv5 }&hard, BoW& inv. index &  \multicolumn{2}{c|}{25k} & -  & - & 73.9 (78.8) &59.3 (65.1)&82.0 (84.8)&0.67kb\\
  && R-MAC \cite{tolias2016particular} &\multicolumn{2}{l|}{vgg16, conv5 }&region MP, SP&PCA$_w$ &-&512& 85.2/86.9 & - & 66.9 (77.3)&61.6 (73.2)&83.0 (86.5)& 1kb\\
  && CroW \cite{kalantidis2016cross} &\multicolumn{2}{l|}{vgg16, conv5}& cross-dim pool. & PCA$_w$ &-&512& -/85.1 & - & 70.8 (74.9)&65.3 (70.6)&79.7 (84.8)& 2kb\\
 && SPoC \cite{babenko2015aggregating} &\multicolumn{2}{l|}{vgg16, conv5}&SP, center prior& PCA$_w$ &-&256& -/80.2 & 3.65 & 53.1$^{\natural}$/58.9&50.1$^{\natural}$/57.8&-& 1kb\\
  && VLAD-CNN \cite{ng2015exploiting} &\multicolumn{2}{l|}{google, various incept.  }&intra-norm, VLAD& PCA&100&128& 83.6 & - & -/55.8&-&-/58.3& 0.5kb\\
\cline{2-15}
   &\multirow{5}{*}{\tabincell{c}{Fine-tuned\\ single-pass} }& \footnotesize Faster R-CNN \cite{salvador2016faster}&\multicolumn{2}{l|}{vgg16-reg-query, conv5 }&SP or MP&-&  -&512 &- & - & 71.0$^{\natural} (78.6)$ &-&79.8$^{\natural} (84.2)$&2kb\\
 && Neural Codes \cite{babenko2014neural} &\multicolumn{3}{l|}{alexnet-classification loss-Landmark, FC6}& PCA& -&128 & -/78.9 & 3.29 & -/55.7 &-/52.3&-&0.5kb\\
  && NetVLAD \cite{arandjelovic2016netvlad}&\multicolumn{3}{l|}{vgg16-trip. loss-Tokyo TM, VLAD layer}&PCA$_w$ & 64&512 &81.7/86.1 & - & 65.6/67.6 &-&73.4/74.9&8kb\\
  && SiaMAC \cite{radenovic2016cnn} &\multicolumn{3}{l|}{vgg16-pair loss-3D Landmark, MAC layer}&L$_w$ & -&512& -/82.5 & 3.61$^{\dag}$ &77.0 (82.9)&69.2 (77.9)&83.8 (85.6)&2kb \\
   && \multirow{1}{*}{Deep Retrieval}&\multicolumn{4}{l|}{vgg16-trip. loss-cleaned Landmark, PCA layer used}&  -&512 &86.7/89.1 & 3.62& 83.1 (89.1) &78.6 (87.3)&87.1 (91.2)&2kb\\
   &&\cite{gordo2016deep,gordo2016end}&\multicolumn{4}{l|}{ResNet101-trip. loss-cleaned Landmark, PCA layer used}&  -&2,048 &90.3/94.8 & 3.84 & 86.1 (90.6) &82.8 (89.4)&94.5 (96.0)&8kb\\
\hline
\end{tabular}
\label{table:compare_methods_datasets}
\end{sidewaystable*}

\textbf{Holidays} \cite{Hamming} is collected by J{\'e}gou \emph{et al.} from personal holiday albums, so most of the images are of various scene types. The database has 1,491 images composed of 500 groups of similar images. Each image group has 1 query, totaling 500 query images. Most SIFT-based methods employ the original images, except \cite{mikulik2010learning,perd2009efficient} which manually rotate the images into upright orientations. Many recent CNN-based methods \cite{babenko2015aggregating}, \cite{arandjelovic2016netvlad}, \cite{kalantidis2016cross} also use the rotated version of Holidays. In Table \ref{table:compare_methods_datasets}, results of both versions of Holidays are shown (separated by ``/''). Rotating the images usually brings 2-3\% mAP improvement.  
\makeatother
\begin{figure*} [t]
\centering
\subfigure[Holidays, \emph{mAP}]{\label{fig:nCam}%
\includegraphics[height=1.8in]{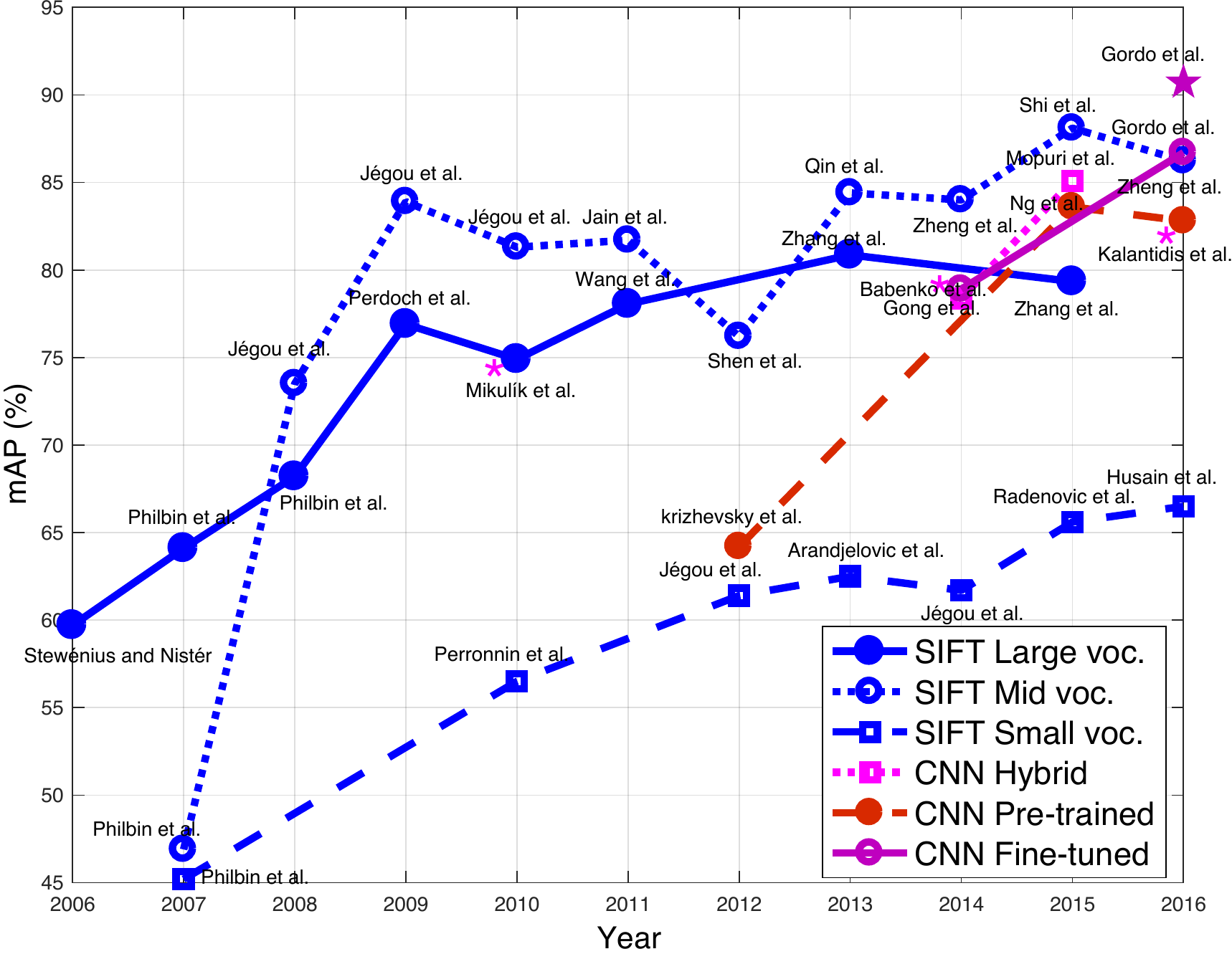}}
 \subfigure[Ukbench, \emph{N-S}]{\label{fig:nFrame}%
\includegraphics[height=1.8in]{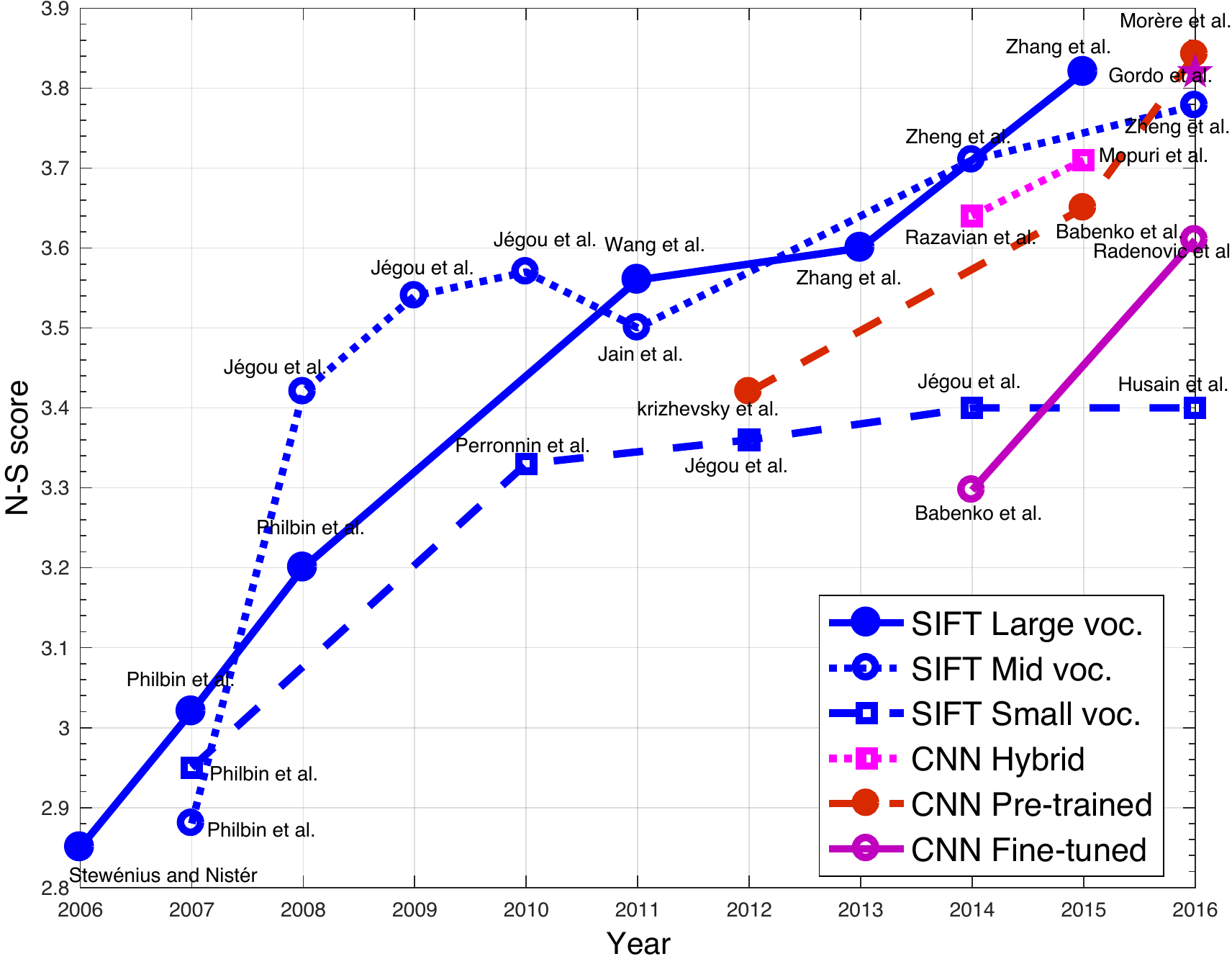}}
 \subfigure[Oxford5k, \emph{mAP}]{\label{fig:nTrack}%
\includegraphics[height=1.8in]{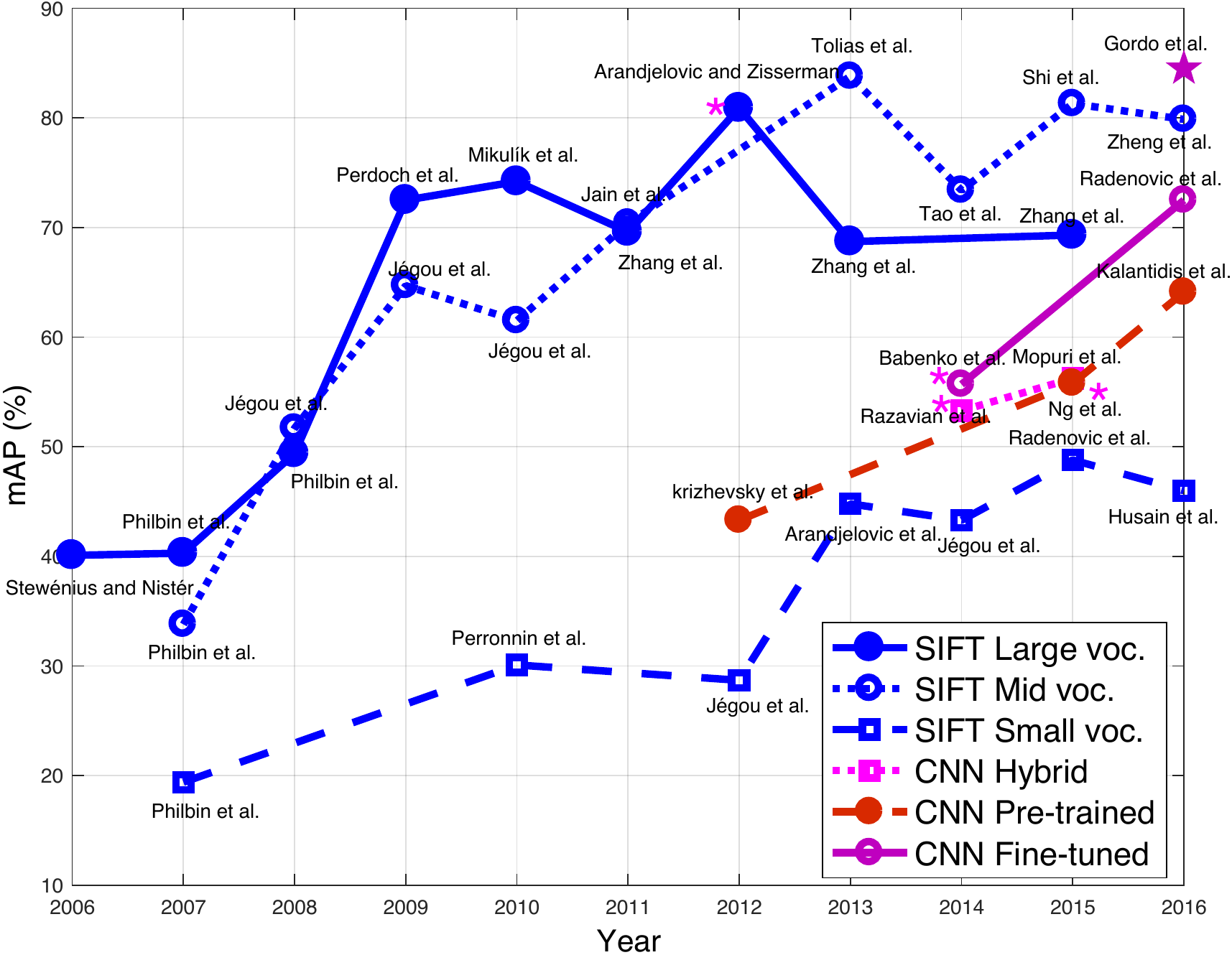}}
\caption{The state of the art over the years on the (a) Holidays, (b) Ukbench, and (c) Oxford5k datasets. Six fine-grained categories are summarized (see Section \ref{sec:retrieval_type}). For each year, the best accuracy of each category is reported. For the compact representations, results of 128-bit vectors are preferentially selected. The purple star denotes the results produced by 2,048-dim vectors \cite{gordo2016deep}, the best performance in fine-tuned CNN methods. Methods with a pink asterisk denote using rotated images on Holidays, full-sized queries on Oxford5k, or spatial verification and QE on Oxford5k (see Table \ref{table:compare_methods_datasets}).}
\label{fig:performance_improvement}
\end {figure*}

\textbf{Ukbench} \cite{HKM} consists of 10,200 images of various content, such as objects, scenes, and CD covers. All the images are divided into 2,550 groups. Each group has four images depicting the same object/scene, under various angles, illuminations, translations, \emph{etc}. Each image in this dataset is taken as the query in turn, so there are 10,200 queries.

\textbf{Oxford5k}  \cite{AKM} is collected by crawling images from Flickr using the names of 11 different landmarks in Oxford. A total of 5,062 images form the image database. The dataset defines five queries for each landmark by hand-drawn bounding boxes, so that 55 query Regions of Interest (ROI) exist in total. Each database image is assigned  one of four labels, \emph{good}, \emph{OK}, \emph{junk}, or \emph{bad}. The first two labels are true matches to the query ROIs, while ``bad'' denotes the distractors. In junk images, less than 25\% of the objects are visible, or they undergo severe occlusion or distortion, so these images have zero impact on retrieval accuracy.

\textbf{Flickr100k}  \cite{soft} contains 99,782 high resolution images crawled from Flickr's 145 most popular tags. In literature, this dataset is typically added to Oxford5k to test the scalability of retrieval algorithms.

\textbf{Paris6k} \cite{soft} is featured by 6,412 images crawled from 11 queries on specific Paris architecture. Each landmark has five queries, so there are also 55 queries with bounding boxes. The database images are annotated with the same four types of labels as Oxford5k. Two major evaluation protocols exist for Oxford5k and Paris6k. For SIFT-based methods, the cropped regions are usually used as query. For CNN-based methods, some employ the full-sized query images \cite{babenko2014neural,arandjelovic2016netvlad}; some follow the standard cropping protocol, either by cropping the ROI and feeding it into CNN \cite{kalantidis2016cross} or extracting CNN features using the full image and selecting those falling in the ROI \cite{salvador2016faster}. Using the full image may lead to mAP improvement. These protocols are used in Table \ref{table:compare_methods_datasets}.


\subsection{Evaluation Metrics}
\textbf{Precision-recall.} Recall denotes the ratio of returned true matches to the total number or true matches in the database, while precision refers to the fraction of true matches in the returned images. Given a subset of $n$ returned images, assuming there are $n_p$ true matches among them, and a total of $N_p$ true matches exist in the whole database, then recall$@n$ ($r@n$) and precision$@n$ ($p@n$) are calculated as $\frac{n_p}{N_p}$ and $\frac{n_p}{n}$, respectively.
In image retrieval, given a query image and its rank list, a precision-recall curve can be drawn on the (precision, recall) points $(r@1, p@1), (r@2, p@2), ..., (r@N, p@N)$, where $N$ is the number of images in the database.

\textbf{Average precision and mean average precision}.
To more clearly record the retrieval performance, average precision (AP) is used, which amounts to the area under the precision-recall curve. Typically, a larger AP means a higher precision-recall curve and thus better retrieval performance. Since retrieval datasets typically have multiple query images, their respective APs are averaged to produce a final  performance evaluation, \emph{i.e.,} the mean average precision (mAP). Conventionally, we use mAP to evaluate retrieval accuracy on the Oxford5k, Paris6k, and Holidays datasets.

\makeatother
\begin{figure*} [t]
\centering
\subfigure[Holidays, \emph{mAP}]{\label{fig:nCam}%
\includegraphics[width=1.927in]{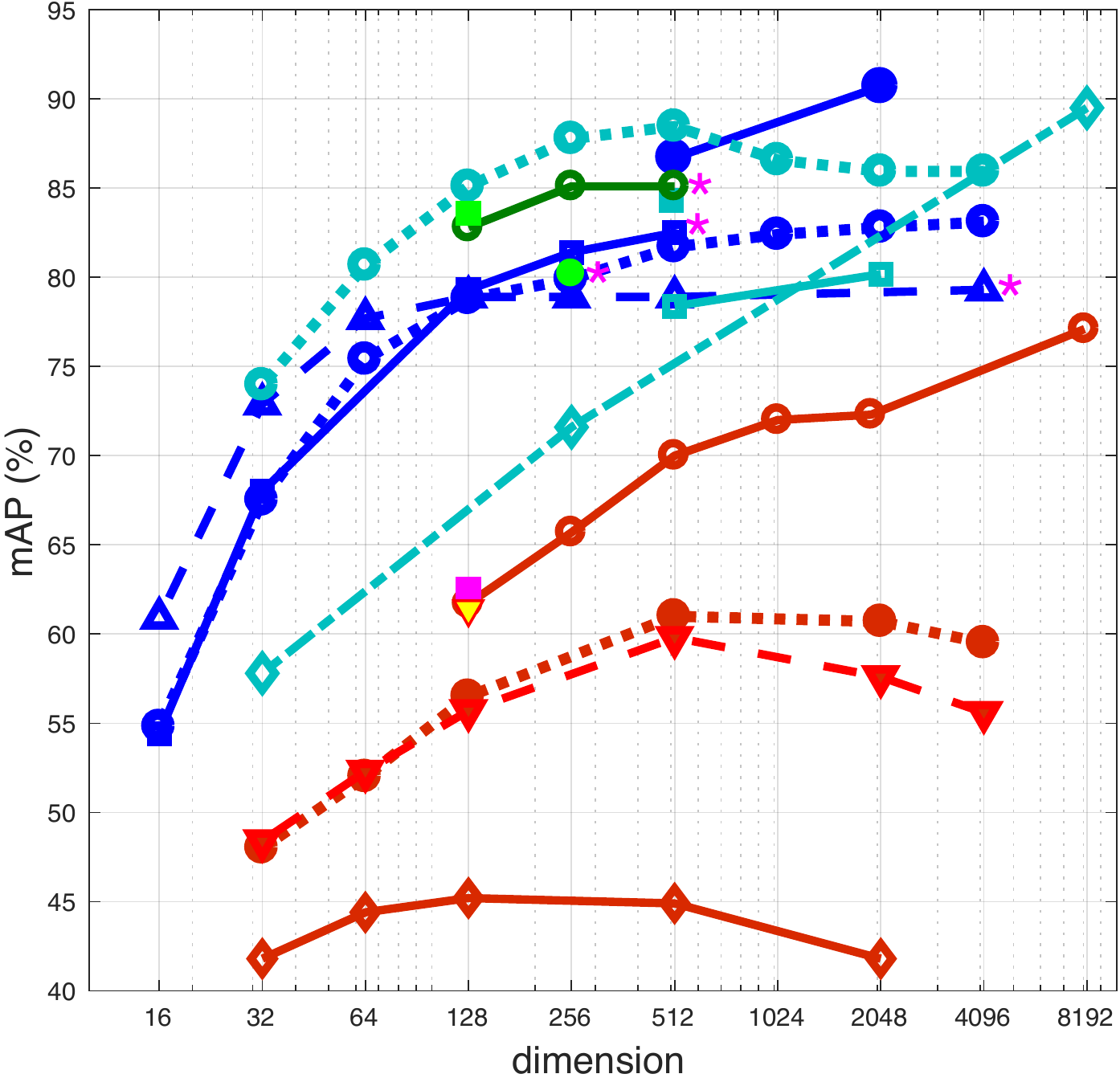}}
\hspace{0.0in}
 \subfigure[Ukbench, \emph{N-S}]{\label{fig:nFrame}%
\includegraphics[width=1.927in]{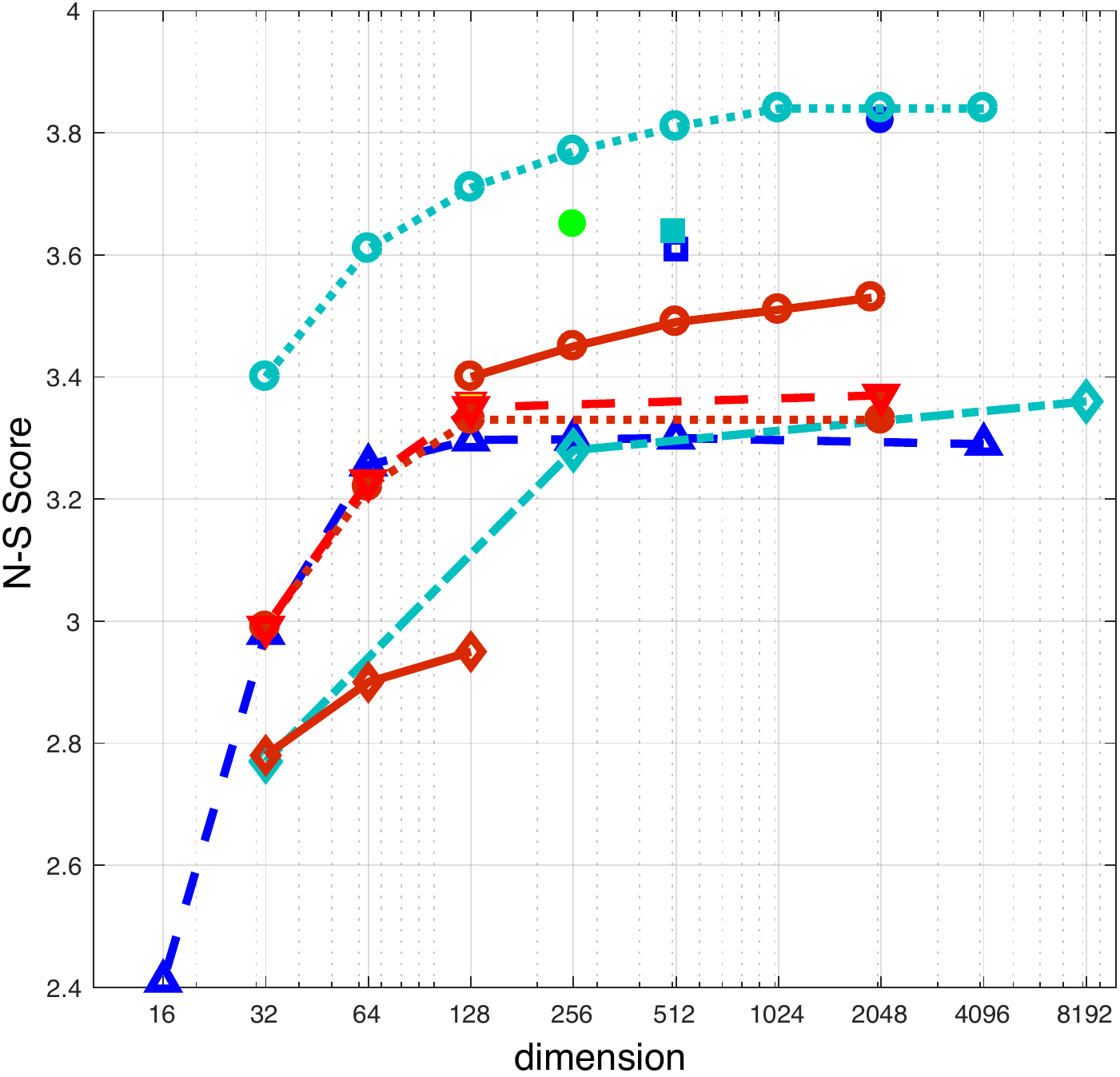}}
\hspace{0.0in}
 \subfigure[Oxford5k, \emph{mAP}]{\label{fig:dim-oxford}%
\includegraphics[width=1.927in]{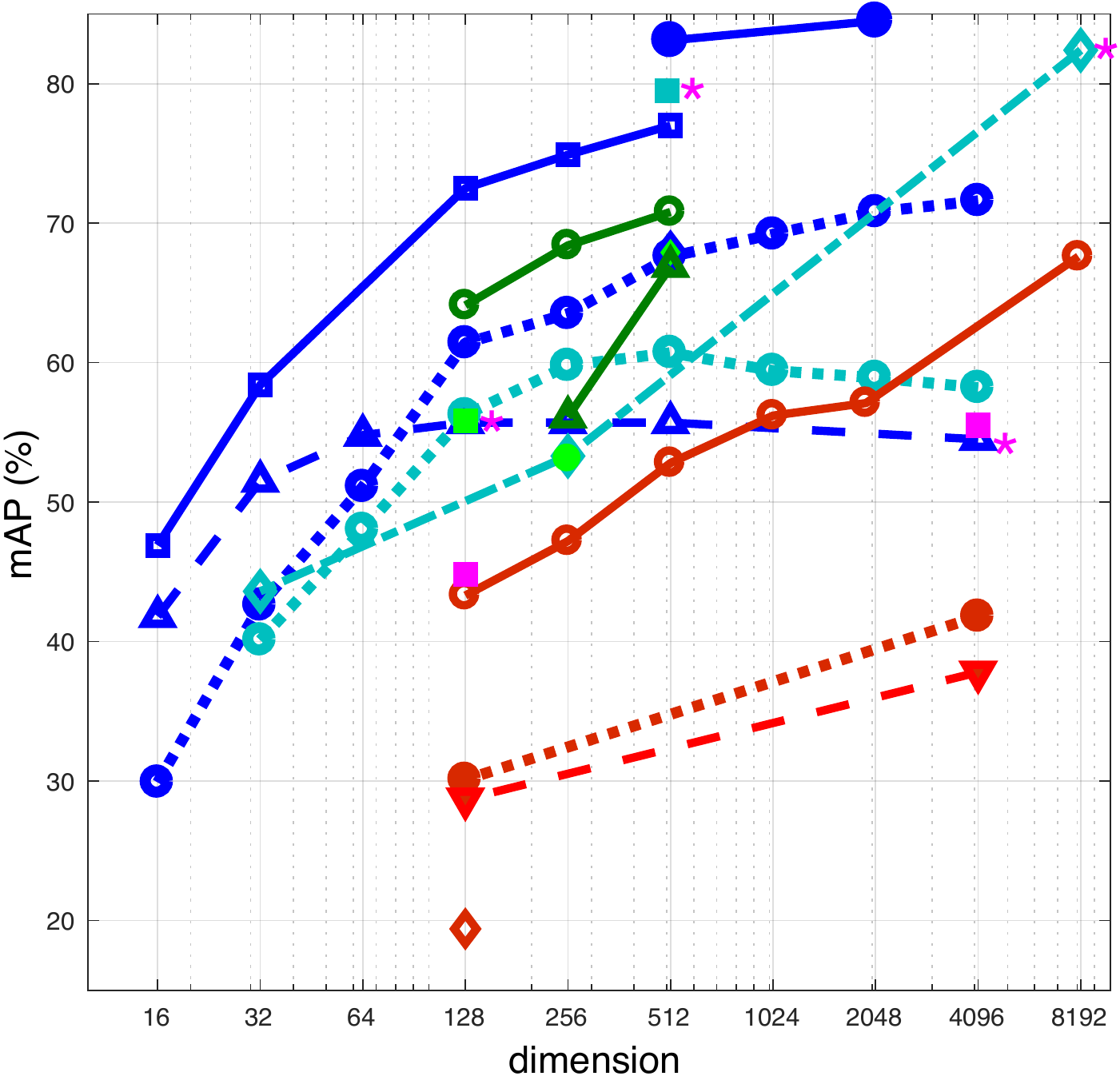}}
\hspace{0.0in}
 \subfigure[Legend]{\label{fig:legend_all}%
\includegraphics[width=1.179in]{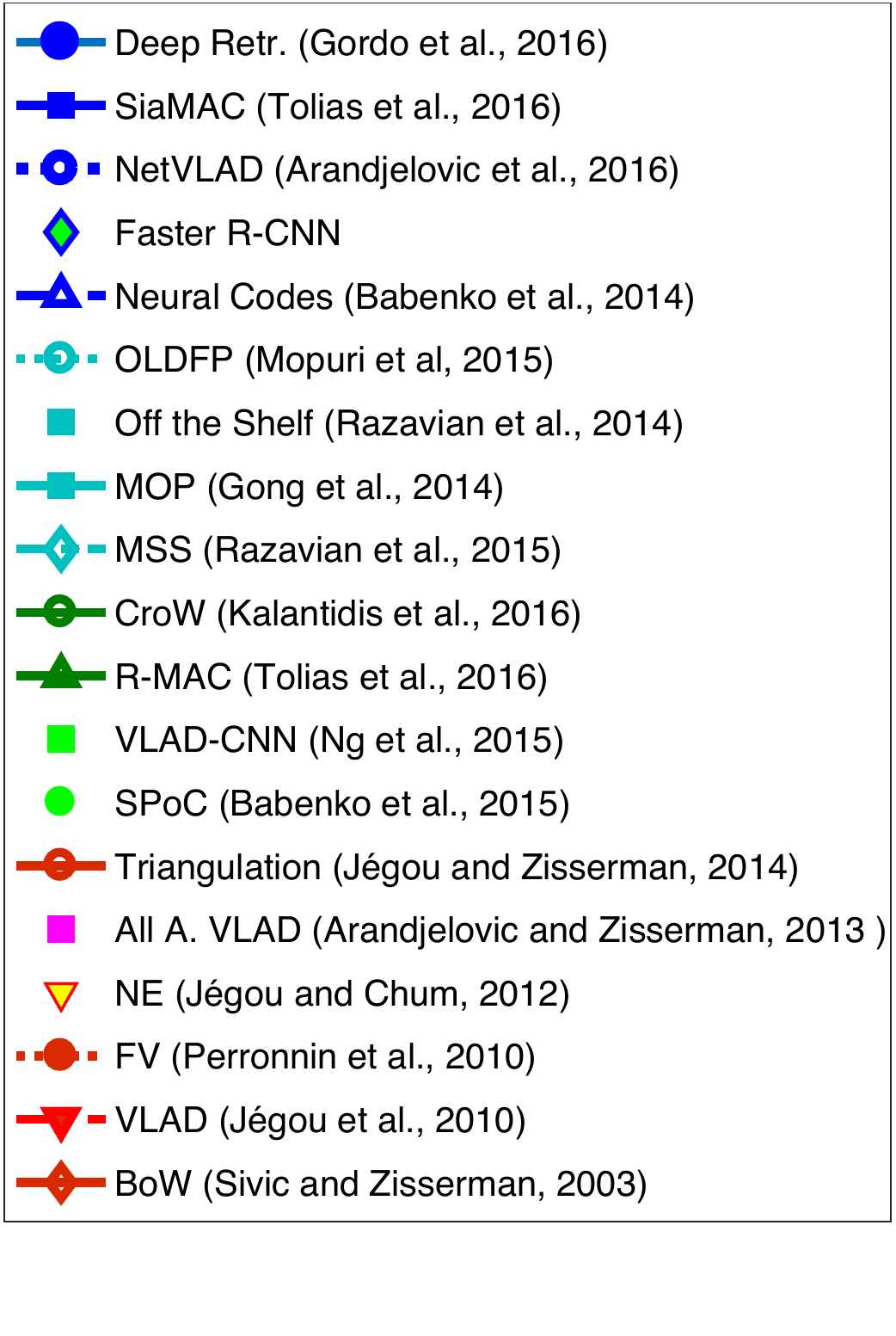}}
\caption{The impact of feature dimension on retrieval accuracy. Compact (fixed-length) representations are shown, \emph{i.e.,} SIFT small voc.,  hybrid CNN methods, pre-trained CNN methods, and fine-tuned CNN methods. Curves with a pink asterisk on the end indicates using rotated images or full-sized queries on Holidays and Oxford5k (see Table \ref{table:compare_methods_datasets}), \emph{resp.}  }
\label{fig:dim_performance}
\end {figure*}
\textbf{N-S Score}. The N-S score is specifically used on the Ukbench dataset and is named after David \textbf{N}ist\'{e}r and Henrik \textbf{S}tew\'{e}nius \cite{HKM}. It is equivalent to precision$@4$ or recall$@4$ because every query in Ukbench has four true matches in the database. The N-S score is calculated as the average number of true matches in the top-4 ranks across all the rank lists.
\subsection{Comparison and Analysis}
\subsubsection{Performance Improvement Over the Years}
We present the improvement in retrieval accuracy over the past ten years in Fig. \ref{fig:performance_improvement} and the numbers of some representative methods in Table \ref{table:compare_methods_datasets}. The results are computed using codebooks trained on independent datasets \cite{Hamming}. We can clearly observe that the field of instance retrieval has constantly been  improving. The baseline approach (HKM) proposed over ten years ago only yields a retrieval accuracy of 59.7\%, 2.85, 44.3\%, 26.6\%, and 46.5\% on Holidays, Ukbench, Oxford5k, Oxford5k+Flickr100k, and Paris6k, respectively. Starting from the baseline approaches \cite{AKM,HKM}, methods using large codebooks improve steadily when more discriminative codebooks \cite{mikulik2010learning}, spatial constraints \cite{GVP,contextual_weighting}, and complementary descriptors \cite{co_indexing,zhang2015cross} are introduced. For medium-sized codebooks, the most significant accuracy advance has been witnessed in the years 2008-2010 with the introduction of Hamming Embedding \cite{Hamming,jegou2010improving} and its improvements \cite{jegou2010improving,burstiness,jain2011asymmetric}. From then on, major improvements come from the strength of feature fusion   \cite{zheng2014packing}, \cite{zhang2015cross}, \cite{zheng2016accurate} with the color and CNN features, especially on the Holidays and Ukbench datasets. 


\setlength{\tabcolsep}{2.6pt}
\begin{table}[t]
\caption{A summary of efficiency and accuracy comparison between different categories. Note: feature extraction time is estimated using CPUs and GPUs (as is usually done) for SIFT and CNN, \emph{resp.}  When using GPUs for SIFT extraction, the efficiency could be high as well.   }
\small
\footnotesize
\centering
\begin{tabular}{l |l l l l | l l}
\hline
\multirow{2}{*}{method type} & \multicolumn{4}{c|}{efficiency} & \multicolumn{2}{c}{accuracy} \\
\cline{2-7}
  & feat. ext. &  retr. & mem.& train & generic & specific \\
\hline
\multirow{1}{*}{SIFT large voc.} &fair& high & fair & fair &high & fair \\
 SIFT mid voc.&fair&low  &low& fair &high&high  \\
 SIFT small voc.&fair & high &high& high &low&Low  \\
\hline
CNN hybrid&low & varies & varies & varies &fair&fair  \\
CNN pre-trained&high & high &high& high &high&fair  \\
CNN fine-tuned &high & high &high& low &high&high  \\

\hline
\end{tabular}
\label{table:complexity_comparison}
\end{table}

On the other hand, CNN-based retrieval models have quickly demonstrated their strengths in instance retrieval. In the year 2012 when the AlexNet \cite{krizhevsky2012imagenet} was introduced, the performance of the off-the-shelf FC features is still far from satisfactory compared with SIFT models during the same period. For example, the FC descriptor of AlexNet pre-trained on ImageNet  yields 64.2\%, 3.42, and 43.3\% in mAP, N-S score, and mAP, respectively, on the Holidays, Ukbench, and Oxford5k datasets. These numbers are lower than \cite{contextual_weighting} by 13.85\%, 0.14 on Holidays and Ukbench, respectively, and lower than \cite{shen2012object} by 31.9\% on Oxford5k. However, with the advance in CNN architectures and fine-tuning strategies, the performance of the CNN-based methods is improving fast, being competitive on the Holidays and Ukbench datasets \cite{gordo2016deep,morere2016group}, and slightly lower on Oxford5k but with much smaller memory cost \cite{radenovic2016cnn}.

\subsubsection{Accuracy Comparisons}
The retrieval accuracy of different categories on different datasets can be viewed in Fig. \ref{fig:performance_improvement}, Table \ref{table:compare_methods_datasets} and Table \ref{table:complexity_comparison}. From these results, we arrive at three observations.

First, among the SIFT-based methods, those with medium-sized codebooks \cite{Hamming,zheng2014packing}, \cite{shi2015early} usually lead to superior (or competitive) performance, while those based on small codebook (compact representations) \cite{jegou2010aggregating,jegou2014triangulation,husain2016improving} exhibit inferior accuracy. On the one hand, the visual words in the medium-sized codebooks lead to relatively high matching recall due to the large Voronoi cells. The further integration of HE methods largely improves the discriminative ability, achieving a desirable trade-off between matching recall and precision. On the other hand, although the visual words in small codebooks have the highest matching recall, their discriminative ability is not significantly improved due to the aggregation procedure and the small dimensionality. So its performance can be compromised.

Second, among the CNN-based categories, the fine-tuned category \cite{babenko2014neural,gordo2016deep,radenovic2016cnn} is advantageous in specific tasks (such as landmark/scene retrieval) which have similar data distribution with the training set. While this observation is within expectation, we find it interesting that the fine-tuned model proposed in \cite{gordo2016deep} yields very competitive performance on generic retrieval (such as Ukbench) which has distinct data distribution with the training set. In fact, Babenko \emph{et al.} \cite{babenko2014neural} show that the CNN features fine-tuned on Landmarks compromise the accuracy on Ukbench. The generalization ability of \cite{gordo2016deep} could be attributed to the effective training of the region proposal network. In comparison, using pre-trained models may exhibit high accuracy on Ukbench, but only yields moderate performance on landmarks. Similarly, the hybrid methods have fair performance on all the tasks, when it may still encounter efficiency problems \cite{xie2015image,razavian2014cnn}.

Third, comparing all the six categories, the ``CNN fine-tuned'' and ``SIFT mid voc.'' categories have the best overall accuracy, while the ``SIFT small voc.'' category has a relatively low accuracy.


\begin{figure}[t]
  \centering
  \includegraphics[width=3.1in]{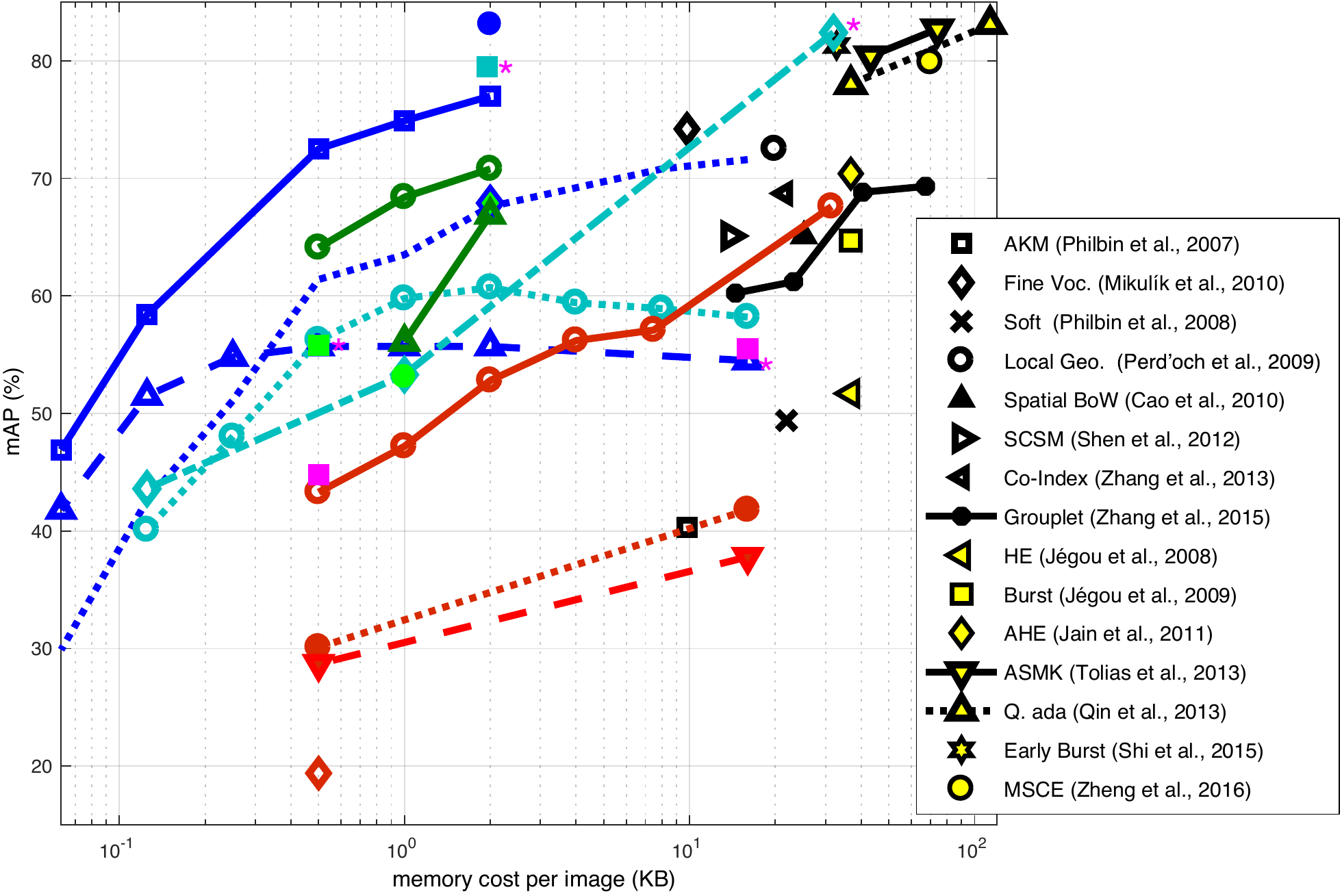}\\
  \caption{Memory cost vs. retrieval accuracy on Oxford5k. In the legend, the first 8 methods are based on large codebooks, while the last 7 use medium-sized codebooks.  This figure shares the same legend with Fig. \ref{fig:dim-oxford} except the newly added numbers (in black).  }\label{fig:mem}
\end{figure}

\subsubsection{Efficiency Comparisons}
\textbf{Feature computation time.} For the SIFT-based methods, the dominating step is local feature extraction. Usually, it takes 1-2s for a CPU to extract the Hessian-Affine region based SIFT descriptors for a 640$\times$480 image, depending on the complexity (texture) of the image. For the CNN-based method, it takes 0.082s and 0.347s for a single forward pass of a 224$\times$224 and 1024$\times$768 image through VGG16 on a TitanX card, respectively. It is reported in \cite{gordo2016deep} that four images (with largest side of 724 pixels) can be processed in 1 second. The encoding (VLAD or FV) time of the pre-trained column features is very fast. For the CNN Hybrid methods, extracting CNN features out of tens of regions may take seconds.  Overall speaking, the CNN pre-trained and fine-tuned models are efficient in feature computation using GPUs. Yet it should be noted that when using GPUs for SIFT extraction, high efficiency could also be achieved.

\textbf{Retrieval time.} The efficiency of nearest neighbor search is high for ``SIFT large voc.'', ``SIFT small voc.'', ``CNN pre-trained'' and ``CNN fine-tuned'', because the inverted lists are short for a properly trained large codebook, and because the latter three have a compact representation to be accelerated by ANN search methods like PQ \cite{jegou2011product}. Efficiency for the medium-sized codebook is low because the inverted list contains more postings compared to a large codebook, and the filtering effect of HE methods can only correct this problem to some extent. The retrieval complexity for hybrid methods, as mentioned in Section \ref{sec:hybrid}, may suffer from the expensive many-to-many matching strategy \cite{razavian2014cnn,razavian2016paper,xie2015image}.

\textbf{Training time.} Training a large or medium-sized codebook usually takes several hours with AKM or HKM. Using small codebooks reduces the codebook training time. For the fine-tuned model, Gordo \emph{et al.} \cite{gordo2016deep} report using five days on a K40 GPU for the triplet-loss model. It may take less time for the siamese \cite{radenovic2016cnn} or the classification models \cite{babenko2014neural}, but should still much longer than SIFT codebook generation. Therefore, in terms of training, those using direct pooling \cite{tolias2016particular,zheng2016good} or small codebooks \cite{jegou2010aggregating}, \cite{ng2015exploiting} are more time efficient.

\textbf{Memory cost.} Table \ref{table:compare_methods_datasets} and Fig. \ref{fig:mem} show that the SIFT methods with large codebooks and the compact representations are both efficient in memory cost. But the compact representations can be compressed into compact codes \cite{jegou2012negative} using PQ or other competing quantization/hashing methods, so their memory consumption can be further reduced. In comparison, the methods using medium-sized codebooks are the most memory-consuming because the binary signatures should be stored in the inverted index. The hybrid methods somehow have mixed memory cost because the many-to-many strategy requires storing a number of region descriptors per image \cite{razavian2014cnn,xie2015image} while some others employ efficient encoding methods \cite{gong2014multi,mopuri2015object}.

\textbf{Spatial verification and query expansion.} Spatial verification which provides refined rank lists is often used in conjunction with QE. The RANSAC verification proposed in \cite{AKM} has a complexity of $\mathcal{O}(z^2)$, where $z$ is the number of matched features. So this method is computationally expensive. The ADV approach \cite{wu2015adaptive} is less expensive with $\mathcal{O}(z\log z)$ complexity due to its ability to avoid unrelated Hough votes. The most efficient methods consist in \cite{avrithis2014hough,schonberger2016vote} which has a complexity of $\mathcal{O}(z)$, and \cite{schonberger2016vote} further outputs the transformation and inliers for QE. 

From the perspective of query expansion, since new queries are issued, search efficiency is compromised. For example, AQE \cite{chum2007total} almost doubles the search time due to the new query. For the recursive AQE and the scale-band recursive QE \cite{chum2007total}, the search time is much longer because several new searches are conducted. For other QE variants \cite{chum2011total}, \cite{root_sift}, the proposed improvements only add marginal cost compared to performing another search, so their complexity is similar to basic QE methods. 

\makeatother
\begin{figure} [t]
\centering
\subfigure[SIFT Large voc.]{\label{fig:nCam}%
\includegraphics[width=1.7in]{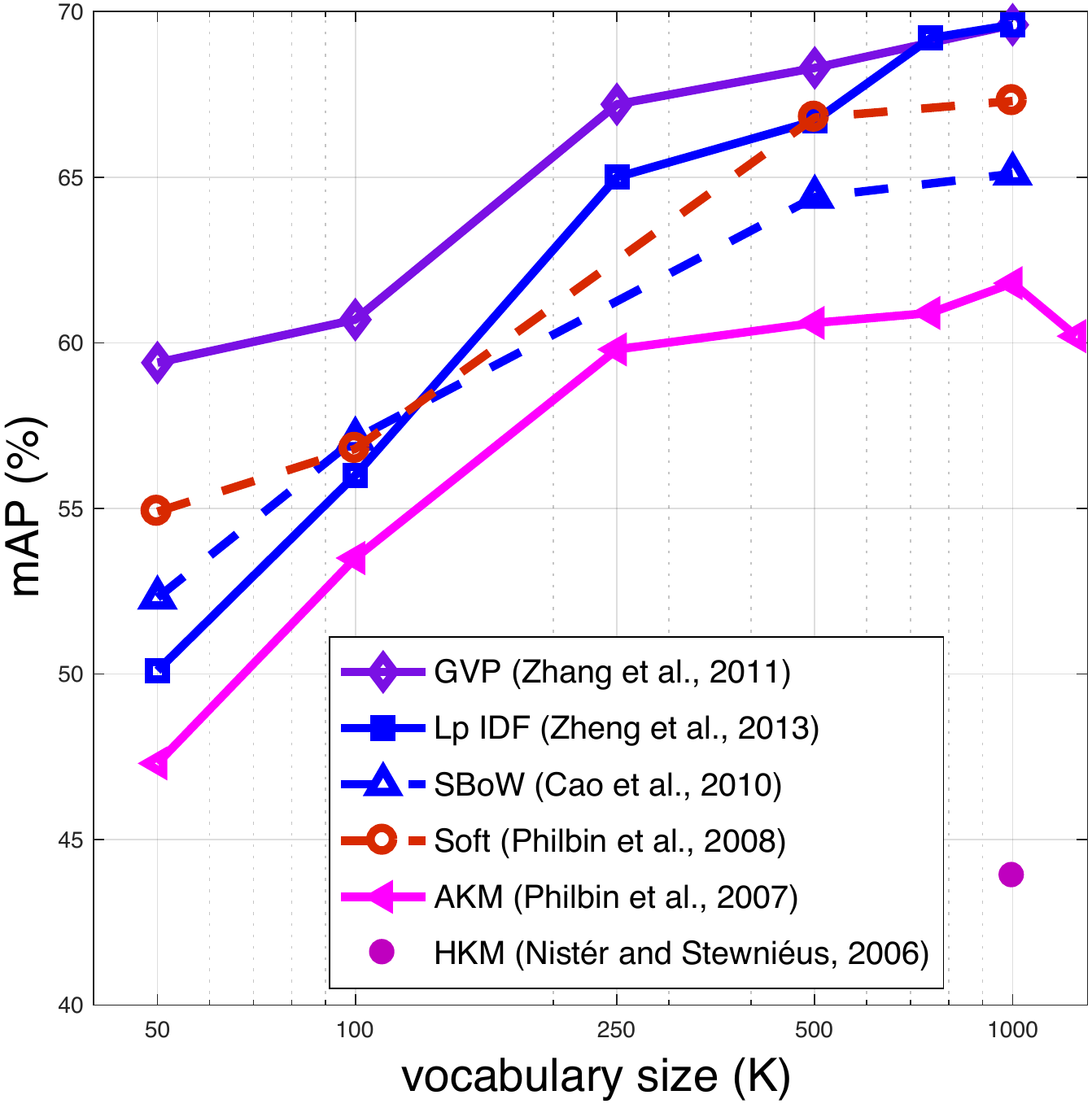}}
\hspace{0.0in}
 \subfigure[SIFT Mid voc.]{\label{fig:nFrame}%
\includegraphics[width=1.7in]{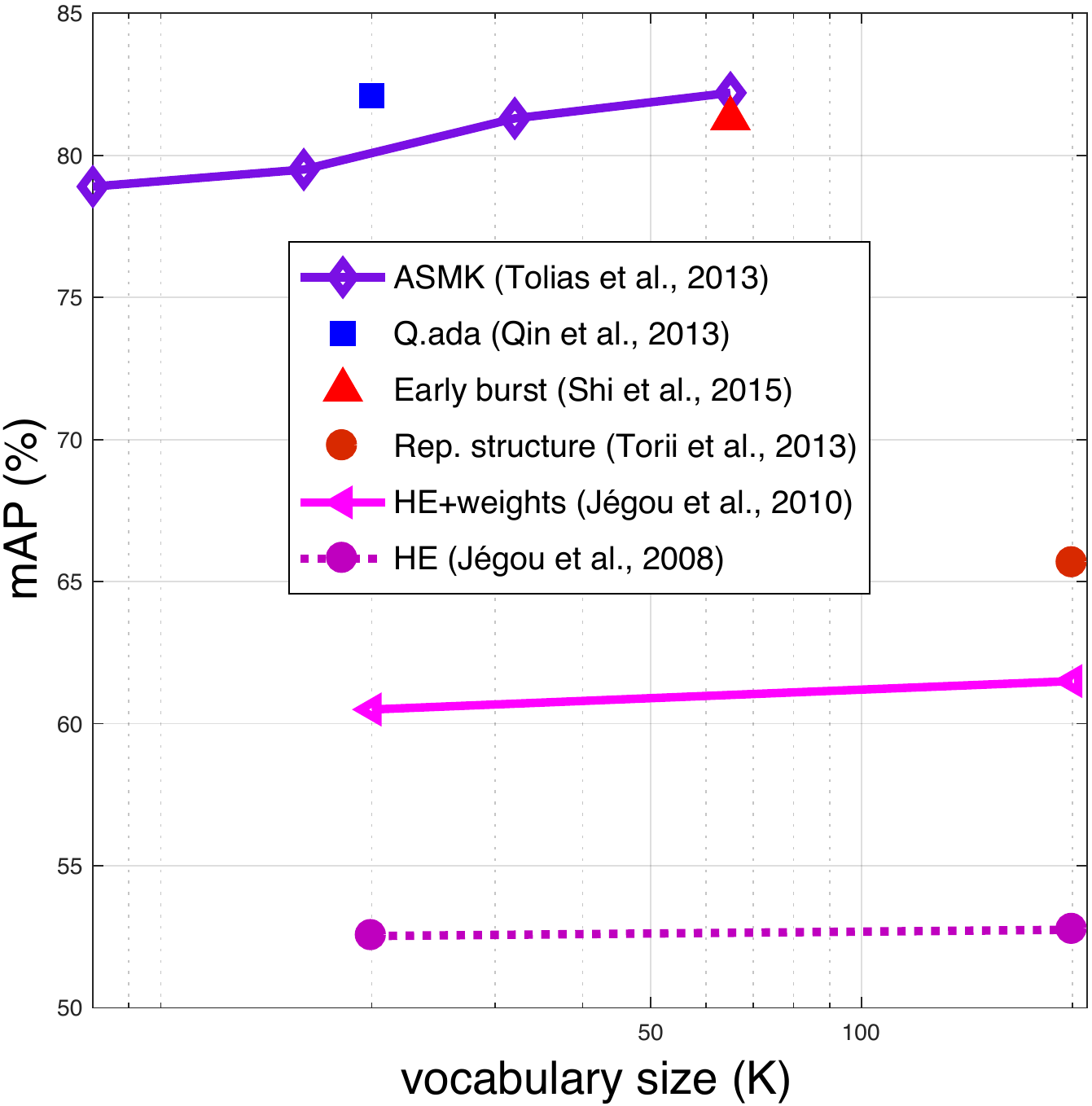}}
\caption{The impact of codebook size on SIFT-based methods using (a) large codebooks and (b) medium-sized codebooks on the Oxford5k dataset. }
\label{fig:voc_performance}
\end {figure}

\subsubsection{Important Parameters}
We summarize the impact of codebook size on SIFT methods using large/medium-sized codebooks, and the impact of dimensionality on compact representations including SIFT small codebooks and CNN-based methods. 

\textbf{Codebook size. } The mAP results on Oxford5k are drawn in Fig. \ref{fig:voc_performance}, and methods using large/medium-sized codebooks are compared. Two observations can be made. First, mAP usually increases with the codebook size but may reach saturation when the codebook is large enough. This is because a larger codebook improves the matching precision, but if it is too large, matching recall is lower, leading to saturated or even compromised performance \cite{AKM}. Second, methods using the medium-sized codebooks have more stable performance when codebook size changes. This can be attributed to HE \cite{Hamming}, which contributes more for a smaller codebook, compensating the lower baseline performance.

\textbf{Dimensionality. } The impact of dimensionality on compact vectors is presented in Fig. \ref{fig:dim_performance}. Our finding is that the retrieval accuracy usually remains stable under larger dimensions, and drops quickly when the dimensionality is below 256 or 128. Our second finding favors the methods based on region proposals \cite{mopuri2015object}, \cite{gordo2016deep}. These methods demonstrate very competitive performance under various feature lengths, probably due to their superior ability in object localization.

\subsubsection{Discussions} 
We provide a brief discussion on when to use CNN over SIFT and the other way around. 
The above discussions provide comparisons between the two features. On the one hand, CNN-based methods with fixed-length representations have advantages in nearly all the benchmarking datasets. Specifically, in two cases, CNN-based methods can be assigned with higher priority. First, for specific object retrieval (\emph{e.g.,} buildings, pedestrians) when sufficient training data is provided, the ability of CNN embedding learning can be fully utilized. Second, for common object retrieval or class retrieval, the pre-trained CNN models are competitive. 

On the other hand, despite the usual advantages of CNN-based methods, we envision that the SIFT feature still has merits in some cases. For example, when the query or some target images are gray-scale, CNN may be less effective than SIFT because SIFT is computed on gray-scale images without resorting to color information. A similar situation involves when object color change is highly intense. In another example, for small object retrieval or when the queried object undergoes severe occlusions, the usage of local features like SIFT is favored. In applications like book/CD cover retrieval, we can also expect good performance out of SIFT due to the rich textures. 

\section{Future Research Directions}\label{sec:directions}


\subsection{Towards Generic Instance Retrieval}
A critical direction is to make the search engine applicable to generic search purpose. Towards this goal, two important issues should be addressed. First, large-scale instance-level datasets are to be introduced. 
While several instance datasets have been released as shown in Table \ref{table:ft_datasets}, these datasets usually contain a particular type of instances such as landmarks or indoor objects. Although the RPN structure used by Gordo \emph{et al.} \cite{gordo2016deep} has proven competitive on Ukbench in addition to the building datasets, it remains unknown if training CNNs on more generic datasets will bring further improvement. Therefore, the community is in great need of large-scale instance-level datasets or efficient methods for generating such a dataset in either a supervised or unsupervised manner.

Second, designing new CNN architectures and learning methods are important in fully exploiting the training data. Previous works employ standard classification \cite{babenko2014neural}, pairwise-loss \cite{radenovic2016cnn} or Triplet-loss \cite{wang2014learning}, \cite{gordo2016deep} CNN models for fine-tuning. The introduction of Faster R-CNN to instance retrieval is a promising starting point towards more accurate object localization \cite{gordo2016deep}. Moreover, transfer learning methods are also important when adopting a fine-tuned model in another retrieval task \cite{bengio2015deep}. 
\subsection{Towards Specialized Instance Retrieval}
To the other end, there are also increasing interests in specialized instance retrieval. Examples include place retrieval \cite{torii201524}, pedestrian retrieval \cite{zheng2015scalable}, vehicle retrieval \cite{liu2016large}, logo retrieval \cite{revaud2012correlation}, \emph{etc}. Images in these tasks have specific prior knowledge that can be made use of. For example in pedestrian retrieval, the recurrent neural network (RNN) can be employed to pool the body part or patch descriptors. In vehicle retrieval, the view information can be inferred during feature learning, and the license plate can also provide critical information when being captured within a short distance.

Meanwhile, the process of training data collection can be further explored. For example, training images of different places can be collected via Google Street View \cite{arandjelovic2016netvlad}. Vehicle images can be accessed either through surveillance videos or internet images. Exploring new learning strategies in these specialized datasets and studying the transfer effect would be interesting. Finally, compact vectors or short codes will also become important in realistic retrieval settings.

\section{Concluding Remarks}\label{sec:conclusion}
This survey reviews instance retrieval approaches based on the SIFT and CNN features. According to the codebook size, we classify the SIFT-based methods into three classes: using large, medium-sized, and small codebook. According to the feature extraction process, the CNN-based methods are categorized into three classes, too: using pre-trained models, fine-tuned models, and hybrid methods. A comprehensive survey of the previous approaches is conducted under each of the defined categories. The category evolution suggests that the hybrid methods are in the transition position between SIFT and CNN-based methods, that compact representations are getting popular, and that instance retrieval is working towards end-to-end feature learning and extraction.


Through the collected experimental results on several benchmark datasets, comparisons are made between the six method categories. Our findings favor the usage of CNN fine-tuning strategy, which yields competitive accuracy on various retrieval tasks and has advantages in efficiency. Future research may  focus on learning more generic feature representations or more specialized retrieval tasks.

\ifCLASSOPTIONcompsoc
  \section*{Acknowledgments}
\else
  \section*{Acknowledgment}
\fi

The authors would like to thank the pioneer researchers in  image retrieval and other related fields. This work was partially supported by the Google Faculty Award and the Data to Decisions Cooperative Research Centre. This work was supported in part to Dr. Qi Tian by ARO grant W911NF-15-1-0290, Faculty Research Gift Awards by NEC Laboratories of America and
Blippar, and National Science Foundation of China (NSFC) 61429201.

\ifCLASSOPTIONcaptionsoff
  \newpage
\fi

\ifCLASSOPTIONcaptionsoff
  \newpage
\fi

\bibliographystyle{IEEEtran}
\bibliography{egbib}

\begin{IEEEbiography}[{\includegraphics[width=1in,height=1.25in,clip,keepaspectratio]{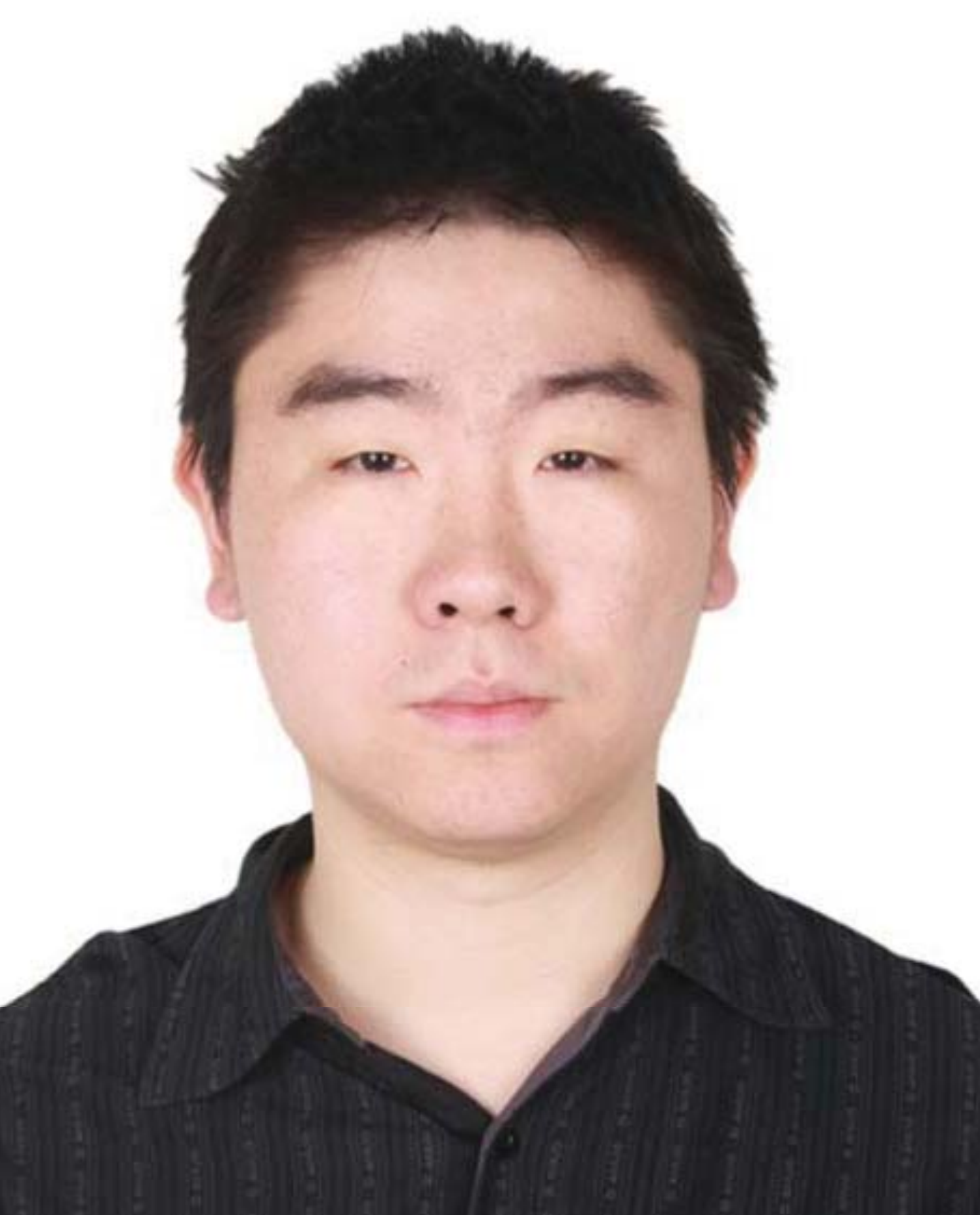}}]{Liang Zheng}
received the Ph.D degree in Electronic Engineering from Tsinghua University, China, in 2015, and the  B.E. degree in Life Science from Tsinghua University, China, in 2010. He was a postdoc researcher in University of Texas at San Antonio, USA. He is now a postdoc researcher in the Center of Artificial Intelligence,  University of Technology Sydney, Australia. His research interests are image retrieval, person re-identification and deep learning.
\end{IEEEbiography}
\begin{IEEEbiography}[{\includegraphics[width=1in,height=1.25in,clip,keepaspectratio]{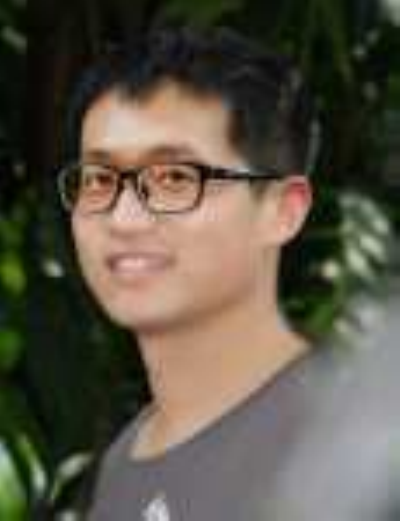}}]{Yi Yang} received the Ph.D. degree in computer
science from Zhejiang University, Hangzhou, China, in 2010. He is currently a Professor with University of Technology Sydney, Australia.
He was a Post-Doctoral Research with the School of Computer Science, Carnegie Mellon University, Pittsburgh, PA, USA. His current research interest includes machine learning and its applications to multimedia content analysis and computer vision.
\end{IEEEbiography}

\begin{IEEEbiography}[{\includegraphics[width=1in,height=1.25in,clip,keepaspectratio]{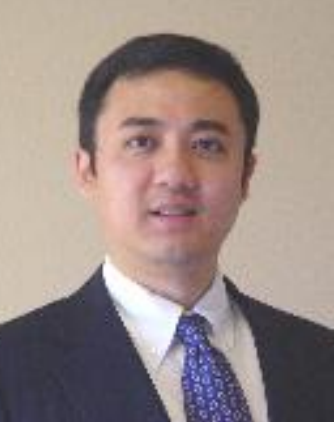}}]{Qi Tian}
(M'96-SM'03-F'16) received the B.E. degree in electronic engineering from Tsinghua University, China, the M.S. degree in electrical and computer engineering from Drexel University, and the Ph.D. degree in electrical and computer engineering from the University of Illinois at Urbana-Champaign, in 1992, 1996, and 2002, respectively. He is currently a Professor with the Department of Computer Science, University of Texas at San Antonio (UTSA). His research interests include multimedia information retrieval and computer vision. \end{IEEEbiography}
\vfill

\end{document}